\documentclass[10pt,journal,compsoc]{IEEEtran}
\ifCLASSOPTIONcompsoc
  \usepackage[nocompress]{cite}
\else
  \usepackage{cite}
\fi

\ifCLASSINFOpdf
   \usepackage[pdftex]{graphicx}
   \DeclareGraphicsExtensions{.pdf,.jpeg,.png}
\else
\fi
\usepackage{times}
\usepackage{epsfig}
\usepackage{graphicx}
\usepackage{amsmath}
\usepackage{amssymb}
\usepackage{booktabs}
\usepackage{color, colortbl}
\usepackage[dvipsnames]{xcolor}

\usepackage[colorlinks,citecolor=red,urlcolor=blue,bookmarks=false,hypertexnames=true]{hyperref}

\usepackage{array}
\newcolumntype{P}[1]{>{\centering\arraybackslash}p{#1}}
\newcolumntype{M}[1]{>{\centering\arraybackslash}m{#1}}

\newcommand{\RemoveBelowCaption}{0pt}

\newcommand{\figref}[1]{Fig.~\ref{#1}}
\newcommand{\secref}[1]{Sec.~\ref{#1}}
\newcommand{\tabref}[1]{Table~\ref{#1}}

\makeatletter
\DeclareRobustCommand\onedot{\futurelet\@let@token\@onedot}
\def\@onedot{\ifx\@let@token.\else.\null\fi\xspace}
\def\eg{\emph{e.g}.} 
\def\ie{\emph{i.e}.}

\def\etal{\emph{et al}.}
\makeatother
\begin{document}

%

\title{Domain-Specific Priors and Meta Learning for \\ Few-Shot First-Person Action Recognition}
%
%
%
%

\author{Huseyin~Coskun,
        M.~Zeeshan~Zia,~\IEEEmembership{Member,~IEEE},
        Bugra~Tekin,
        Federica~Bogo,
        Nassir~Navab,~\IEEEmembership{Fellow,~IEEE},
        Federico~Tombari,
        and~Harpreet~S.~Sawhney,~\IEEEmembership{Fellow,~IEEE}
\IEEEcompsocitemizethanks{\IEEEcompsocthanksitem Huseyin Coskun performed this work during his tenure at Microsoft while he pursued his PhD at Technische Universit\"at M\"unchen.\protect\\
E-mail: huseyin.coskun@tum.de
\IEEEcompsocthanksitem M. Zeeshan Zia is at Retrocausal, Inc. 
\IEEEcompsocthanksitem F. Tombari and N. Navab  are at Technische Universit\"at M\"unchen. 

\IEEEcompsocthanksitem F. Bogo,  B. Tekin, and H. S. Sawhney are at Microsoft.}
}

\IEEEtitleabstractindextext{%
\begin{abstract}
 The lack of large-scale real datasets with annotations makes transfer learning a necessity for video activity understanding. We aim to develop an effective method for few-shot transfer learning for first-person action classification. We leverage independently trained local visual cues to learn representations that can be transferred from a source domain, which provides primitive action labels, to a different target domain -- using only a handful of examples. Visual cues we employ include object-object interactions, hand grasps and motion within regions that are a function of hand locations. We employ a framework based on meta-learning to extract the distinctive and domain invariant components of the deployed visual cues. This enables transfer of action classification models across public datasets captured with diverse scene and action configurations. We present comparative results of our transfer learning methodology and report superior results over state-of-the-art action classification approaches for both inter-class and inter-dataset transfer.
\end{abstract}

}

\maketitle

\IEEEdisplaynontitleabstractindextext

\IEEEpeerreviewmaketitle


\section{Introduction} 
\label{sec:introduction}
Automatically recognizing human hand actions at close range is an important problem for applications such as assembly line inspection, augmented reality training and operations, and egocentric scenarios.
Creating annotated training data for reliable learning of fine scale actions and activities in videos with deep convolutional neural networks is a daunting challenge that limits the scalability, diversity and deployability of research in real world systems. Current approaches either address clip level classification and detection or demand detailed annotations of objects, hands and actions for fine scale recognition.

\begin{figure}[t]
  \centering
    \includegraphics[width=1.0\linewidth]{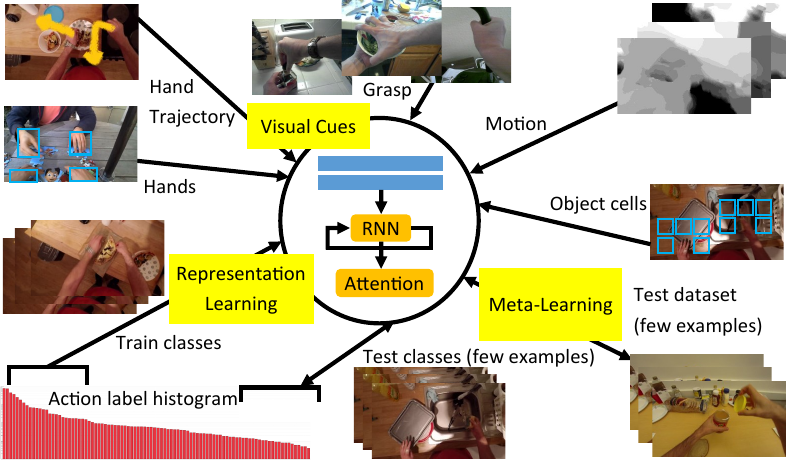}
  \caption{We leverage hand-object-action oriented visual cues in first-person views that decouple foreground action from background appearance, and meta-learning to enable few-shot transfer of action representations. Visual cues are learned from disparate image datasets and applied to first-person action video datasets, thus demonstrating the strength of our decoupled approach.}
 \label{fig:teaser}
\end{figure}

The science of deep video understanding trails behind deep image understanding: for instance, transfer learning, that has been established at least since 2013~\cite{girshick2014rich} for images, was only validated comprehensively in 2017 for videos~\cite{carreira17cvpr} due to the lack 
of large-scale action recognition datasets. Similarly, insights into internal CNN representations that were already available for image-trained CNNs in 
2014~\cite{zeiler2014visualizing} were only comparably studied in 2018~\cite{feichtenhofer2018cvpr} for video CNNs. Likewise, relatively little work exists in learning 
transferable representations for first-person action recognition, and tends to employ relatively complex approaches such as language~\cite{miech17iccv,wray2018cvprw}, sequence alignment~\cite{yang2018cvpr}, or probabilistic reasoning~\cite{rodriguez17cvprw}.

We investigate recognition of primitive actions in first-person videos when only a few training examples are available. We focus on first-person action parsing since it is particularly challenging in its detail and variability of objects and context involved in fine scale actions. First-person action parsing is inherently compositional, in that it involves interaction between hand poses, manipulated objects and motion. Learning truly generalizable deep models for the task requires combinatorially large first-person video datasets. Potentially every object that can be handheld needs to be captured in-situ while covering the entire gamut of hand shapes and appearances, as well as poses and motions that are natural to that specific object and action.

To address the fundamental scalability limitation, we study three important aspects of learning and transfer (see Figure~\ref{fig:teaser}):

1. \textbf{Visual cue learning from independent datasets}: We propose using cues relevant to first-person action understanding as strong priors to regularize learning and inference. It is unrealistic to assume availability of labels for every visual cue relevant to a given video understanding task in a single training set. We propose a set of a priori visual cues for action recognition with the idea that these cues can be learned using disparate datasets without the need for all the associated annotations in a single action dataset. We test our approach on first-person action datasets where the types of annotations available are limited. For instance, numerous datasets are available for object detection: we investigate the efficacy of transferring this training to action datasets while using object level context and detection.
   
2. \textbf{Inter-class transfer}: We observe that most existing first-person action datasets are severely imbalanced~\cite{damen2018eccv,li2015cvpr} with few actions represented by a large number of examples while most actions fall in the long tail of the action distribution. We investigate training action classification models on one set of classes with relatively large number of training examples and transferring to (testing on) a disjoint set of classes given only a few training examples for the latter. Success in this endeavor would enable incrementally adding actions ``on-the-fly" to the set of recognizable actions without demanding a large number of annotations. 
    
3. \textbf{Inter-dataset transfer}: We also explore transfer learning as per its most common interpretation, \emph{i.e.} by transferring models trained on one dataset to another dataset.
 Achieving this capability would enable models that can generalize across significant scene and object appearance variations, \emph{e.g.} allowing fine tuning a model originally trained for ``kitchen activity'' on factory floor tasks.

To tackle few-shot learning problems, we explore the use of meta learning for action recognition.
Recent work~\cite{finn2017model,finn2018probabilistic,ravi2017} successfully applied meta learning to image classification tasks; however, its use for video classification has received less attention~\cite{finn2017model}. We build on the Model-Agnostic Meta-Learning (MAML) algorithm~\cite{finn2017model}, combining it with an attention mechanism to improve its performance on temporal sequences. We call this approach Attentive MAML (A-MAML).

We validate a complete pipeline (Section~\ref{sec:approach}) for transferring first-person action representations across datasets and classes. Ablation studies yield insights into the nuances of performing such transfer (Section~\ref{sec:experiments}). Figure~\ref{fig:teaser} depicts our use of multiple contextual cues trained on their respective datasets and used in our investigation to demonstrate multiple transfer learning tasks.

Our contributions are summarized as follows:
\begin{enumerate}
    \item
    We introduce strong priors for object and action context, that decouple foreground action from background appearance in first-person videos. Object and action context includes hand regions as hard focus-of-attention cues, grasp classification, class-agnostic object-object interactions, and hand trajectory. {\em We demonstrate successful transfer of these image-only cues, derived from diverse image-only datasets, to video action recognition}. To the best of our knowledge, we are the first to combine multiple hand-crafted cues in the context of few-shot learning, for first-person action recognition.
    \item We demonstrate the effectiveness of these 
    domain cues in the context of transfer learning, for both inter-class and inter-dataset transfer. We perform a thorough evaluation across two large-scale first-person action recognition datasets, outperforming both significant ablative baselines as well as state-of-the-art action classification approaches.
    \item  We propose Attentive MAML, which combines MAML~\cite{finn2017model} with an attention mechanism for effective training for few-shot learning.
\end{enumerate}

The structure of the paper is as follows: Sect.~\ref{sec:related} surveys both classical and recent literature on action recognition, transfer learning, as well as first-person action scenarios. Sect.~\ref{sec:approach} details our system and its components, followed by Sect.~\ref{sec:experiments} which introduces our experimental protocol and results of our investigations. Finally, Sect.~\ref{sec:conclusion} concludes the paper summarizing lessons learned.
\section{Related Work}
\label{sec:related}
We review below the rich literature on action recognition, transfer learning and few-shot learning, with a particular focus on first-person actions.

\textbf{Hand-crafted features for action recognition.} Early approaches in third-person scenarios (\emph{e.g.} surveillance videos) typically rely on hand-crafted spatio-temporal features such as HOG-3D~\cite{klaser08}, STIP~\cite{laptev05}, SIFT-3D~\cite{scovanner2007} or dense trajectories~\cite{matikainen09iccvw} and combine them using bag-of-words. Egocentric videos present specific challenges, like camera motion, large occlusions, background clutter~\cite{li2015cvpr}. In this scenario, traditional visual features have been shown to perform poorly~\cite{fathi11,fathi13,pirsiavash12}. A number of approaches proposed to focus instead on object-centric representations~\cite{fathi11,fathi11cvpr,pirsiavash12}. Additionally, the use of egocentric cues like camera and head motion~\cite{kitani11,li13,ryoo13}, hand motion and pose~\cite{li13} and gaze information~\cite{fathi12,li13} has been explored. Li~\etal ~\cite{li2015cvpr} provide a systematic analysis of motion, object and egocentric cues for first-person action recognition, showing that they can be combined with motion-compensated traditional features.

\textbf{Deep  features  for  action  recognition.} With the growing availability of data~\cite{heilbron15,carreira17cvpr,sigurdsson16,ucf101} and the advent of deep learning, the emphasis has shifted towards learning video features with deep neural networks (DNNs). DNN architectures for action recognition can be roughly grouped into three classes: 3D ConvNets, Recurrent Neural Networks (RNNs) and two-stream networks.
3D ConvNets~\cite{ji13,karpathy14,taylor10,tran15,varol18} extend standard convolutional networks to deal with the temporal dimension by using spatio-temporal filters. While they can learn hierarchical representations of spatio-temporal data, their high number of parameters makes training hard and data-hungry~\cite{carreira17cvpr}. RNN-based approaches rely on 2D ConvNets, usually pre-trained for image classification, to extract features from each frame in a video, and then capture temporal structure by adding a recurrent layer (\emph{e.g.} an LSTM) to the model~\cite{donahue15,ng15}. LSTMs have been used in conjunction with attention mechanisms~\cite{li18,sharma16}. Two-stream architectures~\cite{feichtenhofer16,simonyan14,wang16} rely on two parallel 2D ConvNets -- the so called ``spatial'' and ``flow'' streams -- usually obtained from pre-trained image classification models. The spatial stream takes as input RGB frames; the flow stream takes as input pre-computed optical flow fields. Action predictions are obtained by averaging the output from the two streams. Recently, Carreira and Zisserman~\cite{carreira17cvpr} proposed Two-Stream Inflated 3D ConvNets (I3D), which model spatial and flow streams with two 3D ConvNets.

Working with long videos has been shown particularly challenging, since models need to capture long-range relationships between frames. To this end ~\cite{girdhar2017actionvlad,hussein2019timeception,wu2019long,hussein2019videograph,lu2019learning,luo2019grouped} propose to use either multi-scale temporal convolutions or attention mechanisms to capture relationship within long range videos.

\textbf{Deep  features  for first-person  action recognition.} Similar ideas have been applied to egocentric scenarios. However, progress has been limited by the lack of huge amounts of annotated data, which only recently started to become available~\cite{damen2018eccv,Goyal2017,li13}. Ma~\etal ~\cite{ma16} fuse an appearance and a motion stream in a multi-task architecture to predict primitive actions, objects and activities. They explicitly train the appearance stream to learn egocentric features like hand segmentation and object labels. Similarly, other approaches explicitly focus on hand- or object-based features. Bambach~\etal~\cite{Bambach_2015_ICCV} train a CNN to segment hands, and argue that hand segments alone can be used to recognize actions. Cai~\etal~\cite{Cai16} recognize grasp type, object attributes and action class in a unified framework. The recently proposed Object Relation Network (ORN)~\cite{Baradel_2018_ECCV} classifies activities by learning contextual relationships between detected semantic object instances. In all these works, it remains unclear how well these learned features generalize across datasets.

\textbf{Transfer learning.} There exists a rich literature~\cite{cook13} on transfer learning of ``pre-CNN'' features for action recognition, focusing especially on transfer across input modalities. In videos, Karpathy~\etal~\cite{karpathy14} transfer CNN-based features learned on a huge dataset of sport videos to recognize actions on UCF-101~\cite{ucf101}. Sigurdsson~\etal~\cite{sigurdsson18} learn a joint representation for actions from first- and third-person videos. In the egocentric domain, Wray~\etal~\cite{wray2018cvprw} propose to map a single action to multiple verb labels to generalize video retrieval across datasets, though they do not provide quantitative results. To the best of our knowledge, no work has provided an analysis of transfer learning methodologies for egocentric action recognition so far.

\textbf{Few-shot learning.} While few-shot learning has been actively studied for image understanding purposes~\cite{andrychowicz2016,finn2017model,li2017,ravi2017,vinyals2016}, it has received far less attention for video understanding ~\cite{yang2018cvpr,BishayZP19,zhu2018compound,zhang2020few}. Recently proposed methods for video-based few-shot learning combine deep metric learning~\cite{vinyals2016} with episode-based training strategies~\cite{snell2017prototypical,vinyals2016}. While metric learning approaches measure similarity to few-shot example inputs, episode-based training strategies aim to learn how to update the parameters for a small number of examples.  Along this line of research, TARN~\cite{BishayZP19} proposes a metric learning technique that consists of an embedding and a relation module; the proposed model is trained with an episode-based training strategy. While the embedding module is responsible to embed samples to encode, the relation module aligns the encoded samples and learns a deep metric on them. TARN classifies samples based on the average distance to labeled samples for each class. Similarly to TARN, Zhang \etal\cite{zhang2020few} propose a deep metric learning method taking aligned video clips as input. Using metric learning, Compound Memory Networks (CMN)~\cite{zhu2018compound} propose a multi-saliency embedding algorithm to obtain fixed-size features from videos of arbitrary length. This embedding is then used for video retrieval and classification. Metric-learning based approaches rely on algorithms like the K-Nearest Neighbors (KNN) to  classify samples, so the number of labeled samples available per class significantly impacts the performance. As a consequence, such approaches tend to have generalization issues when there are quite a few samples per class.

\begin{figure*}
  \centering
    \includegraphics[clip, trim=4cm 11cm 4cm 10cm, width=0.95\textwidth]{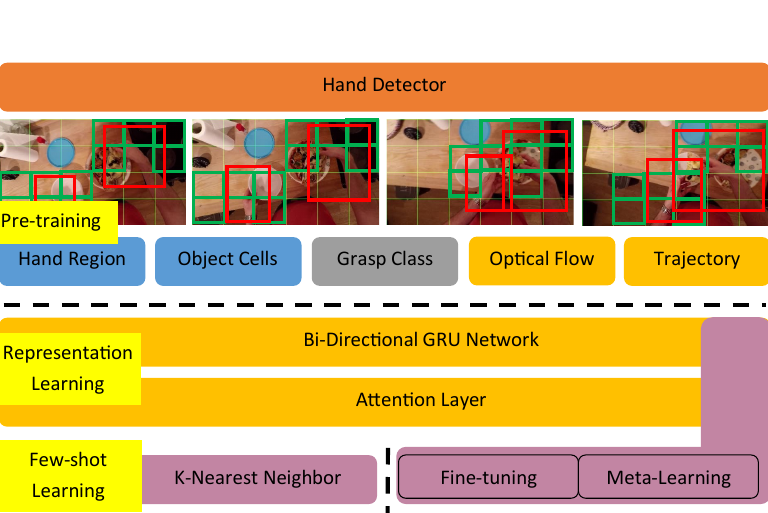}
  \caption{System overview: The hand detector provides extended \emph{hand-context} (red boxes) and activates object cells (dark green) laid out in a fixed grid (light green). 
  Training proceeds in three stages: (i) visual cue extractors, trained from disparate image datasets, (ii) Bi-Directional GRU and Attention layer, trained from source domain videos with action labels, and (iii) transfer learning: we experiment with fine-tuning, KNN and meta learning. }
 \label{fig:pipeline}
\end{figure*}
\section{Approach} \label{sec:approach}

\begin{figure}
  \centering
    \includegraphics[clip,trim=1cm 3cm 2cm 3cm,width=0.48 \textwidth]{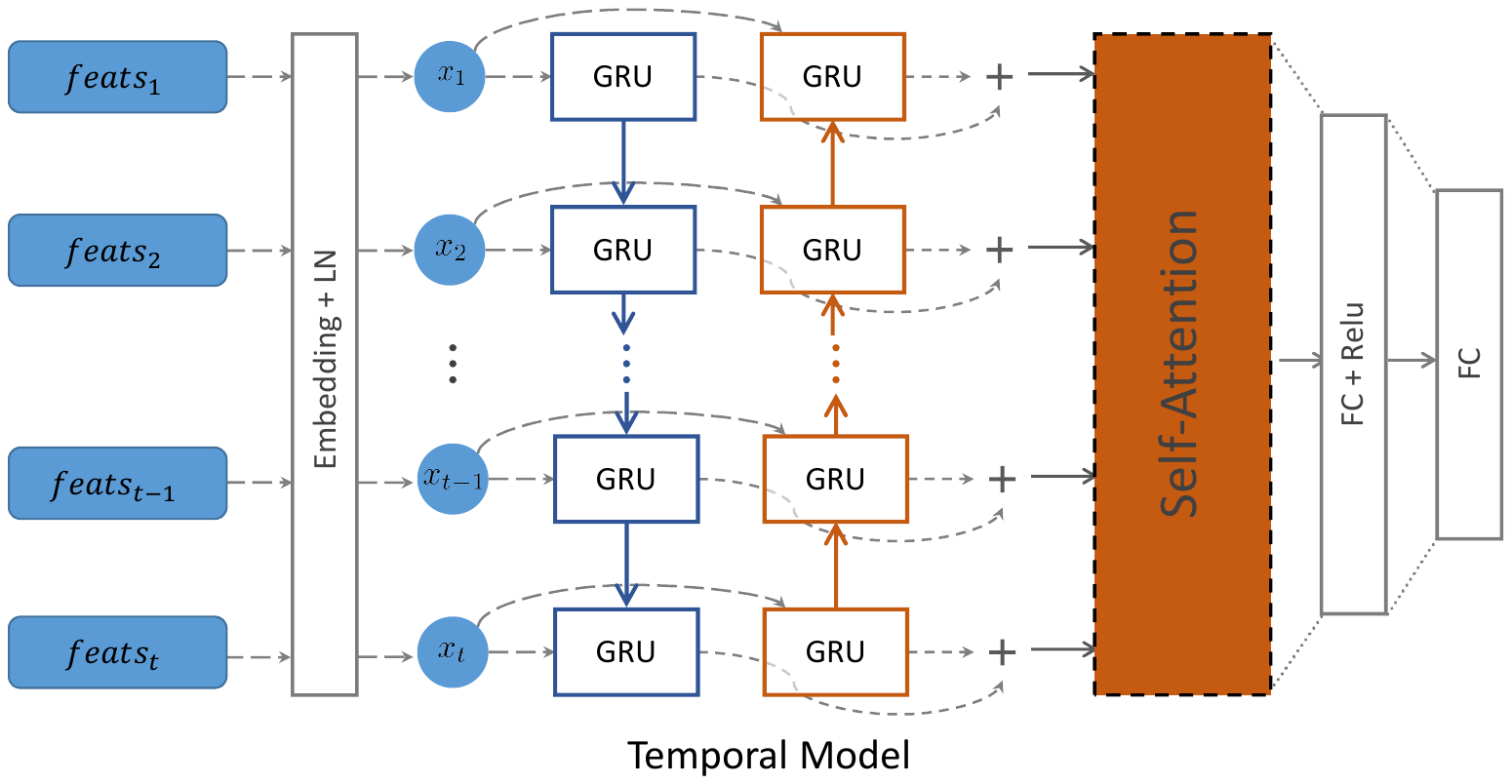}
  \caption{Architecture overview: Feats are $1$-d vectors and, depending on the experiment, are a  concatenation of different visual cues. FC represents a fully connected layer; LN represents layer normalization. The embedding layer ensures that the number of parameters of the temporal model stays fixed, even if different visual cues are used.}
 \label{fig:tm}
\end{figure}
We conjecture that a key problem in transferring learned representations for activity understanding in first-person videos is the strong coupling between hand-object interaction (``the action'') in the foreground, and the appearance of
the background. Since datasets for first-person action are still fairly small, training an action recognition model on a given dataset inevitably causes the model to over-fit to coincidental regularities in scene appearance.

We aim to generalize action learning by focusing the learned model on the foreground activity while neglecting background appearance. Knowledge of typical first-person action configuration provides us with an opportunity to inject inductive biases such as the existence of left and right hands, hand-object configurations and motion. We exploit these to decouple foreground action context from the background. Specifically, we extract multiple features (``visual cues'') inside the hand region, as obtained from a hand bounding box detector, and use them to train temporal models from a source dataset that is different from the target action dataset.

Since most action recognition datasets do not contain all the relevant visual cue labels, we utilize cue-relevant image datasets to train feature representations for the visual cues. Figure~\ref{fig:pipeline} depicts our use of hand regions and object context regions within a multi-layer architecture. Off-the-shelf hand detection is used to define regions occupied by hands, as well as object regions (called ``object cells") near the hands. We also compute grasp features using public domain pre-trained network features, and motion cues such as optical flow and pixel trajectories in the hand regions. Each of these high-dimensional features is embedded in a learned fixed dimensional representation that is used to train a bi-directional temporal network whose output is processed by an attention layer. The features for hand and object regions are computed using a pre-trained object detection network and grasp gestures use a pre-trained grasp network. 

\subsection{Visual Cues}
\label{sec:visual_cues}

We define a-priori visual cues and process them through their respective DNNs to compute feature descriptors as a pre-cursor to training a temporal model. More details about the feature extraction process are provided in Table~\ref{tbl:cues} and Section ~\ref{sec:experiments}.

\textbf{Hand detection as ``hard attention mechanism''. }
We train a Faster R-CNN~\cite{ren2015nips} detector to localize the left and right egocentric hands in single video frames. We assume that the left hand always appears in the left half of the egocentric frame, and right hand in the right half; so we extract only the highest scoring bounding boxes (above a threshold score) in the left and right halves of the image.
We observed that this assumption is valid for most of the frames, given the egocentric viewpoint. We further employ a tracker to predict hand location from previous detections if the detector fails to provide a bounding box. In turn, we compute CNN-derived features for visual cues from local regions defined around the left and right hand bounding boxes. 

\textbf{Local hand context.} 
Intuitively, first-person actions are implied by the region of the scene close to the subject's hands, including any objects being held in, or close to the hands~\cite{sudhakaran2019lsta,kumar2015fly}. We extend the bounding boxes by a fixed factor $s_{hand}$, such that it includes local context, which we refer to as ``hand-context box'' (red bounding boxes in Figure~\ref{fig:pipeline}). We compute \emph{mid-level} convolutional features from a pre-trained object detector by Region-of-Interest (RoI) pooling~\cite{girshick2015fast} within the extended bounding boxes.

\textbf{Object-interaction modeling.} In a transfer learning setting, we cannot rely on the knowledge of explicit object classes for hand-object and object-object reasoning~\cite{Baradel_2018_ECCV}. A universal action model should be able to represent 
unknown object classes. A model trained on, say, a ``kitchen'' dataset should transfer its detection capability, using few-shot learning, to test videos captured in a factory environment, showing different, unknown tools.

We draw inspiration from~\cite{santoro2017nips} in which a simple grid cell representation of the scene to model object-object interactions is applied on the ``CLEVR'' visual question-answering dataset~\cite{johnson2017cvpr}. We define a fixed grid over a frame, and ``activate'' the five cells surrounding the hand detection bounding box (green boxes in Figure~\ref{fig:pipeline}). The object features are computed as the \emph{max-pool} operation (element-wise max) over 
features obtained per activated cell. We interpret this operation as computing a feature descriptor representing object-object and hand-object interactions. 

\textbf{Grasp features.} Visual hand pose estimation is a challenging problem, which becomes even harder when hands manipulate objects. However, we do not need precise joint location to correctly interpret an interaction. With this intuition, we draw inspiration from the robotic manipulation literature, and utilize fine-grained grasp classification~\cite{rogez2015understanding} as an approximation of hand poses, to aid action classification. Specifically, we train a CNN on cropped \emph{hand-context} regions to predict grasp classes~\cite{rogez2015understanding}, and use the pre-logit feature vector as our grasp features.

\textbf{Optical flow features.} We follow common knowledge from the action recognition literature~\cite{davis97cvpr,simonyan14} to inject explicit motion cues as training data. 
Specifically, we modify an Inception-ResNet-v2 network~\cite{szegedy2017inception} to feed in the horizontal and vertical optical flow channels for \emph{K} frames centered around the
current frame. The network is trained to classify the optical flow frames into action classes \emph{w.r.t.} training labels. Again, we extract mid-level convolutional features~\cite{girshick2015fast} within the \emph{hand-context box} as the feature representation which gets fed into the temporal model. Please see \tabref{tbl:training_details} for more details of training and dataset.

\textbf{Hand trajectory features.} In addition to using optical flow to explicitly represent motion, we also capture longer term motions within the context of the moving hands. We encode image trajectories of interest points~\cite{matikainen09iccvw} to aid action understanding. We use the 2D bounding box coordinates of the hands over the past few frames as our 
trajectory feature.

\subsection{Network Architecture for Action Training} \label{sec:net_arch}
We utilize a standard recurrent neural network (RNN) architecture as the backbone for temporal modeling.

\textbf{Temporal modeling.} Figure~\ref{fig:tm} shows our model. The temporal model consists of an embedding layer followed by a bi-directional GRU network~\cite{chung2014empirical}. The embedding layer comprises a fully connected layer of size N $\times$ $256$ and layer normalization, where \emph{N} denotes the total dimensionality of all the visual cue input features. Our bi-directional GRU has $128$ hidden units. We use the same architecture for all experiments, adapting the input dimensionality by the visual cue size, for the sake of comparing various cue-combinations.

\textbf{Attention layers.} We employ the recently proposed self-attention mechanism of \cite{lin2017structured} to process the outputs of the GRU network.
In particular, given a sequence of frame features $X=\{x_1, x_2,  \dots, x_{n}\}$ with a length of $n$, we first compute GRU outputs ${\bf S} =\{s_{1}, s_{2}, \dots, s_{n}\}$, $S \in R^{n \times 2u}$, where $u$ represents the number of hidden units at GRU and $s_{t}$ is the concatenation of forward and backward GRU state vectors at time $t$. Formally we can write it as in the following: 
\begin{align}
s_{t} = [\overrightarrow{GRU}(x_{t},s_{t-1}),\overleftarrow{GRU}(x_{t},s_{t+1})]
\end{align} 
where $x_{t} \in R^u$ is the frame features at time $t$. Note that $s_{t}$ contains information which is computed via information flow in forward and backward directions in time. After obtaining the state matrix $S$, we then compute the attention matrix ${\bf A}  \in R^{r \times n}$. Formally: 
\begin{align}
A = softmax(M)
~~~~\text{and}~~~~
{\bf M} ={\bf W_{s2}}\tanh({\bf W_{s1}} {\bf S}^\top) 
 \label{eq:attn}
\end{align} 
%
%
where ${\bf W_{s1}} \in R^{h \times 2u}$ and ${\bf W_{s2}} \in R^{r \times h}$ denote weight matrices. $h$ and $r$ are hyperparameters that represent the number of hidden units and the number of attention heads, respectively. Each attention head enables the model to focus on different parts of the video. 
We choose $r=3$ and $h=100$ based on cross-validation and we use these values for all the experiments. After obtaining the attention matrix $A$, we can use it to embed the given state matrix into a fixed-size vector: 

\begin{align*}
&E=AS, \hspace{3mm} E \in ^{r \times 2u}   
\end{align*}

\noindent Finally, we flatten the matrix $E$ into a vector to obtain our video representation. Note that the embedding dimension is independent of the size of the sequence and allows us to represent videos of arbitrary size in a fixed-sized vector. As in~\cite{lin2017structured}, a naive implementation of this attention mechanism results in a model in which all attention heads are focusing on the same frame at time $t$ in the sequence. To overcome this issue, we use a regularization term to enforce the attention matrix $A$ to focus on different parts of the sequence.
Lin~\etal~\cite{lin2017structured} show that the dot product of the attention weight matrix and its transpose, subtracted by the identity matrix, can be used as a measure of diversity:

\begin{align}
P =|| {\bf A} {\bf A^T} - {\bf I}||_F  \label{eq:attn-norm}
\end{align} 
where ${\bf I}$ is the identity matrix and $F$ indicates the Frobenius norm. We add the matrix $P$ to the loss function during training to encourage the diversity of attention weight vectors. 

\subsection{Training for Transfer}
\label{sec:train_transf}

We hypothesize that the visual cues introduced in Sec.~\ref{sec:visual_cues} are particularly effective for transfer learning. We test the hypothesis using three different methods for few-shot learning and by testing on unseen classes/datasets.
Firstly, we employ K-Nearest Neighbors (KNN) on the learned attention features. Secondly, we use fine-tuning of the action recognition network for few-shot learning on a few target domain examples by transferring the RNN representation learned on the source datasets. Thirdly, we employ an attentive meta-learning mechanism building on~\cite{finn2017model}. 

\textbf{KNN.} We use features extracted from the attention layer of our network in a KNN framework to perform few-shot matching. Specifically, we randomly select \emph{L} samples per test class and classify an activity via majority vote of its neighbors. Model prediction is assigned to the most frequent label among its \emph{K} nearest neighbors. For evaluation, we repeat this process $50$ times and average the results. 

\textbf{Fine-tuning.} We select \emph{L} test samples per class and fine-tune our model using these sequences. We perform two different types of fine-tuning: only logits and full parameters. In our experiments all models are trained for the same number of updates. We select $L$ sequences per class and fine-tune our models using only these sequences. We fine tune each model 15 times.
Each time, we train the model for 200, 2k and 10k iterations, setting $L = 1$, $L = 10$, and $L = 30$, respectively. The logit layer parameters are initialized uniformly at random. We repeat the fine tuning experiments 15 times for each model and report mean and standard deviation values of the model accuracy.

\textbf{Attentive Model Agnostic Meta Learning (A-MAML).} We build on the MAML training algorithm proposed in~\cite{finn2017model}, which aims at finding an optimal set of network parameters that can be transferred to new classes using only a few training samples and updates.
The algorithm consists of two parts -- \emph{meta-learner} and \emph{learner}. While the \emph{meta-learner} trains over a range of tasks, the individual learners optimize for specific tasks. Here, the tasks correspond to $K$-shot $N$-class action classification problems. In the course of the algorithm, first, a support set consisting of $K$ labelled examples for $N$ classes is selected from the full training data. In the context of few-shot learning, $K$ corresponds to a small number of training examples. The learner optimizes over this training set for a small number of updates. While specializing on a specific task, the network does not yet generalize across a variety of different tasks. We would like to find a set of optimal network parameters such that the overall model generalizes across tasks and task-specific learning requires fewer updates. To this end, a meta-objective is defined across different tasks minimizing the same training loss. We sample a new task at the end of each training and update the objective function with gradient descent on the new support set.

Experimentally, we observed convergence issues when training our network with MAML. As also reported by~\cite{antoniou2019}, MAML is difficult to train; the use of RNNs exacerbates this difficulty~\cite{martens2011learning,pascanu2013difficulty}, since its optimization relies on vanilla gradient descent during \emph{learner} updates. We overcome this problem by optimizing the task specific loss \emph{only} over the attention layer parameters. Attention has been shown to be more effective than RNNs for transfer learning~\cite{radford2018improving}, but not within a meta-learning context. We demonstrate that attention-based meta-learning results in improved accuracy and higher generalization power for few-shot action recognition.  In this framework, the bidirectional GRU acts as a class-agnostic embedding network, while the attention layer effectively acts as the classifier. 
\section{Experimental Results}\label{sec:experiments}
\begin{table*}[!t]
\centering
\scalebox{0.9}{
\begin{tabular}
{
			>{\raggedright\arraybackslash}p{2.5cm}|
			>{\centering\arraybackslash}p{3cm}|
			>{\centering\arraybackslash}p{1.0cm}|
			>{\centering\arraybackslash}p{2.7cm}|
			>{\centering\arraybackslash}p{2.2cm}|
			>{\centering\arraybackslash}p{2.0cm}|
			>{\centering\arraybackslash}p{1.7cm}} 
\hline
Network & Opt & LR & LR Decay & Opt Params & Batch Sz. & \#Iteration  \\
\hline
Flow Network & Adam& 0.001 & Exp Decay (0.95) & Adam-Default & 10 & 100K \\
Grasp Network & Adam& 0.001 & Exp Decay (0.95) & Adam-Default & 10 &   10K \\
Hand Detector & Adam& 0.001 & Exp Decay (0.95) & Adam-Default & 4 & 50K \\
I3D~\cite{carreira17cvpr}  & Adam& 0.01 & Exp Decay (0.95) & Adam-Default & 10 & 150K \\
TRN~\cite{zhou2018temporal} & SGD + Momentum& 0.001 & Each 50K ($\times 0.5$) & Momentum(0.9) & 10 & 150K \\

Temporal Model  & Adam& 0.001 & Exp Decay (0.95) & Adam-Default & 10 & 100K \\

\hline
\end{tabular}}	
\vspace{5mm}
        \caption{Visual cue networks training details.} 
		\label{tbl:training_details}
\end{table*}

\begin{table*}[!t]
\centering
\scalebox{1}{
\begin{tabular}
{
			>{\raggedright\arraybackslash}p{1.9cm}|
			>{\centering\arraybackslash}p{3.0cm}|
			>{\centering\arraybackslash}p{1.5cm}|
			>{\centering\arraybackslash}p{5.5cm}|
			>{\centering\arraybackslash}p{2.7cm}|
			>{\centering\arraybackslash}p{1.0cm}} 
\hline
Cue & Network & Feature Layer Used & Training Task & Training Dataset & Dim \\
\hline
Coco-Global & Inc-ResNet-v2& Mixed\_6a & Object Detection (Faster R-CNN) & COCO &   1088 \\
Flow-Global & Inc-ResNet-v2& Mixed\_6a & Activity Classification& EPIC (12 Classes) &  1088 \\
Hand & Inc-ResNet-v2& Mixed\_6a & Object Detection (Faster R-CNN) & COCO & 1088*2 \\
Obj & Inc-ResNet-v2& Mixed\_6a & Object Detection (Faster R-CNN ) & COCO&  1088*2 \\
Flow & Inc-ResNet-v2& Mixed\_6a & Activity Classification & EPIC (12 Classes) & 1088*2 \\
Grasp & Inc-ResNet-v2& Mixed\_6a & Grasp Classification & GUN71 & 1088*2  \\
Traj & Inc-ResNet-v2& Final & Hand Detection (Faster R-CNN) & EgoHands  & 10*4*2  \\
\hline
\end{tabular}}	
\vspace{5mm}
        \caption{Network-specific details for domain-specific visual cue extraction. Note that we have visual cues for left and right separately  hence the 2 multipliers for the visual cues except for global cues. \emph{Mixed\_6a} layer is the one just before the \emph{PreAuxLogits} layer, in the  Inception-ResNet-v2~\cite{szegedy2017inception} base network.} 
		\label{tbl:cues}
		
\end{table*}

\textbf{Datasets.} We evaluate our model on two relatively large first-person video datasets: 
EPIC Kitchens (EPIC)~\cite{damen2018eccv} and Extended GTEA Gaze+ (EGTEA)~\cite{li13}. \textbf{EPIC} contains $55$ hours of recordings, featuring $28$ subjects performing daily activities in different kitchen scenarios. Videos are recorded by a GoPro camera at  $1920 \times 1080$ resolution. Each subject records and labels their own activities. The dataset provides labels as primitive actions, \emph{i.e.} verbs that define the action of the subject (\emph{e.g.} close, open, wash). In total there are $125$ such primitive action labels. \textbf{EGTEA} also records activities in kitchen scenarios and contains $106$ different activity videos from  32 subjects. Each subject prepares a complete dish. In total, EGTEA provides $28$ hours of recordings. Videos are labeled by means of (\emph{verb}, \emph{noun}) pairs. To keep label similarity with EPIC we consider only verb labels. This gives us 19 distinct primitive activity labels. EGTEA is recorded with an SMI wearable eye-tracker with a resolution of  $1280 \times 960$. 

In order to evaluate the ability of our models on \emph{inter-class} transfer learning, we define a specific training/test split on EPIC that highlights the typical data distribution skews in that a few classes are well represented while others are sparse. However the results are typically reported on the whole ensemble. We choose well-represented frequent action classes with samples for training, and less represented classes for testing. This methodology can be applied to almost all the datasets in the computer vision literature and would reveal nuances of various approaches.
Our training set includes frames from 12 classes (\emph{close}, \emph{cut}, \emph{mix}, \emph{move}, \emph{open}, \emph{pour}, \emph{put}, \emph{remove}, \emph{take}, \emph{throw}, \emph{turn-on}, \emph{wash}). For each of these classes, we have more than 300 sequences in EPIC.
Note that these correspond to fairly generic actions, hence they are good candidates for training a base network.
While there are fewer sequences for our test classes in EPIC, given the long tail distribution of activities in the dataset they provide a representative test set (we have on average $20$ sequences per test class).
We use 5 subjects (\emph{S01}, \emph{S10}, \emph{S16}, \emph{S23}, \emph{S30}) for testing and the rest of the subjects for training. This split ensures that training and test sets have the same number of classes for a fair experiment.
We train our flow and grasp networks, the hand detector and the temporal model using parameters detailed in Table~\ref{tbl:training_details}. The Table also details the parameters we use for training four state-of-the-art baselines for action classification: I3D~\cite{carreira17cvpr}, Two Stream I3D~\cite{carreira17cvpr}, TRN~\cite{zhou2018temporal}, and Two Stream TRN~\cite{zhou2018temporal}.

\definecolor{Res6}{rgb}{0.93,0.93,0.93}
\definecolor{Res5}{rgb}{0.99,0.94,0.91}
\definecolor{Res3}{rgb}{0.93,0.96,0.91}
\definecolor{Res4}{rgb}{0.93,0.95,0.98}
\definecolor{Res2}{rgb}{0.95,0.95,0.95}
\definecolor{Res1}{rgb}{1.0,0.97,0.90}

\begin{table}[t]
	\centering
	\begin{tabular}{
	>{\raggedright\arraybackslash}p{2.75cm}|
	>{\centering\arraybackslash}p{1.4cm}|
	>{\centering\arraybackslash}p{1.4cm}} 
		\midrule		
		Rogez~\etal \cite{rogez2015understanding}& 22.67\\
		Ours & \textbf{42.32}\\
		\bottomrule
	\end{tabular}
	\setlength{\belowcaptionskip}{\RemoveBelowCaption}
	\vspace{4mm}
	\caption{Grasp accuracy (in \%) on the GUN71 dataset~\cite{rogez2015understanding}.} \label{tbl:gun71_grasp}
\end{table}

\subsection{Training Visual Cue, Temporal and Baseline Models} \label{sec:visual-cue-training}
In Table~\ref{tbl:training_details} we provide details about the training parameters used for the various visual cues employed in our approach. Details of each will be highlighted in the text below.

In Table~\ref{tbl:cues}, we present an overview of the network specific details for extraction of various domain-specific visual cues. In the following, we specify the datasets used to learn each of the visual cues introduced in Sec.~\ref{sec:visual_cues} and briefly describe the training procedures adopted for our RNN model, and for the baselines. Note that all the cue extractor networks are based on an Inception-Resnet-v2 \cite{szegedy2017inception} backbone.

\noindent \textbf{Hand detector.} We use the EgoHands dataset~\cite{Bambach_2015_ICCV} to train our hand detector. The dataset is recorded with Google Glass and contains complex, first-person interactions between two people and most of the images include two or more hands. We adopt the \emph{Labeled Data} split provided by the authors. This split contains $4,800$ images. The dataset provides ground-truth hand masks. We compute bounding boxes corresponding to the ground truth hand regions, and use this as training data. Our hand detector employs the atrous version of Faster R-CNN~\cite{girshick2015fast}. We initialize the network weights with a model pre-trained on COCO~\cite{coco}. More details about training our hand detector can be seen in \tabref{tbl:training_details}.

\begin{figure}[t]
	\centering
	\includegraphics[clip, trim=0cm 2cm 0cm 0cm,width=1\linewidth]{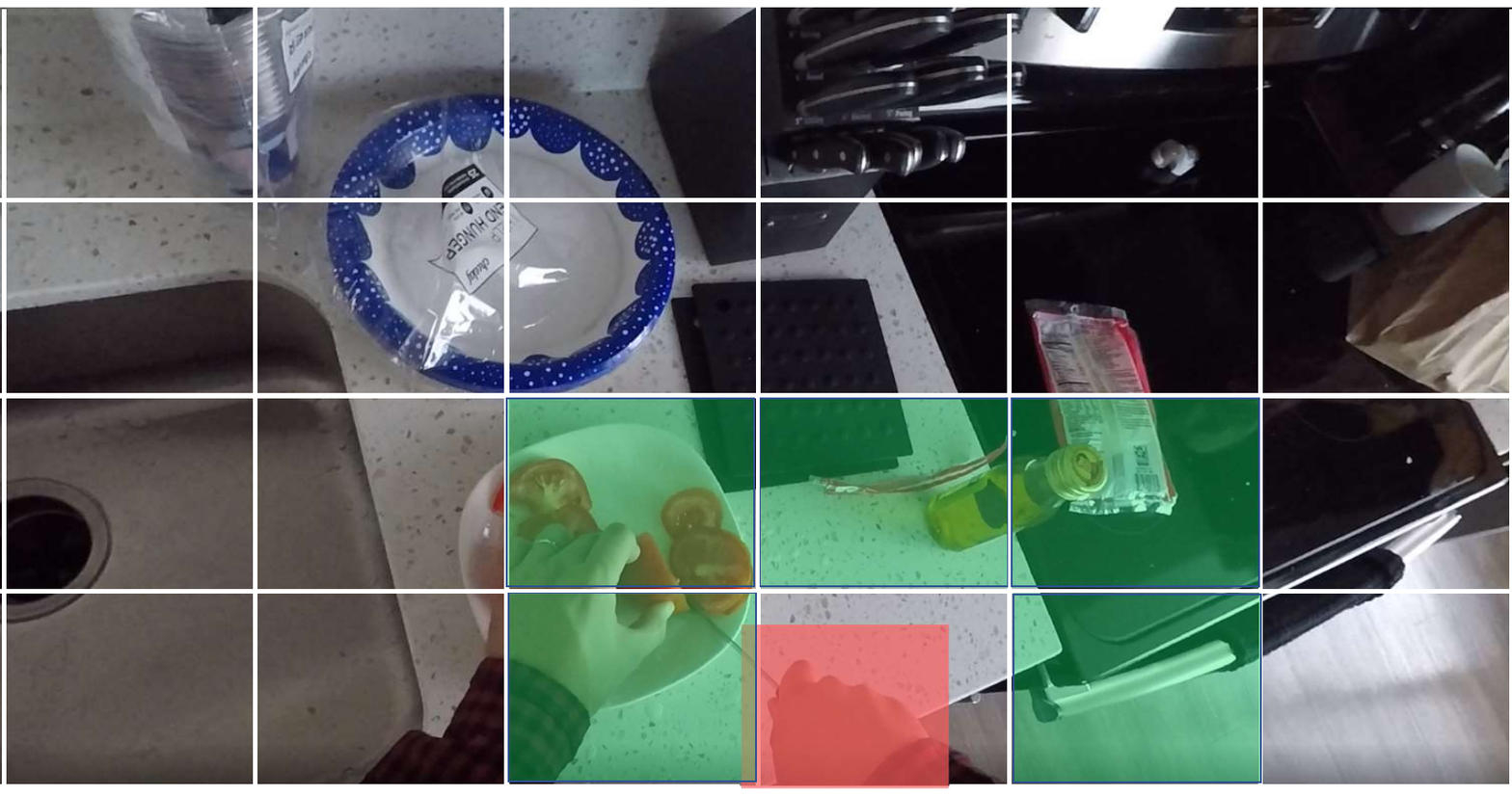}
	\caption{We visualize the right-hand feature extraction regions in image space. We subdivide the image into $4 \times 6$ cells. Green cells show our object feature extraction regions. The red box shows the hand feature extraction region. The changes in the red box size depend on the hand size in the image.
	}	
	\label{fig:feature_regions}
\end{figure}

\noindent \textbf{Hand-context, object cell and global features.} Figure \ref{fig:feature_regions} shows our hand and object cell regions for one hand. 
We extract feature descriptors for hand context and object cell regions from an Inception-ResNet-v2 (Faster R-CNN backbone, \emph{Mixed\_6a} features), trained for MS-COCO object detection.
\emph{Hand} features are obtained by max pooling over the hand region only. For \emph{object} features we subdivide the input image into 4 $\times$ 6 cells and extract features from the cells surrounding the hand center cells (left, top-left, top, top-right, and right).
\emph{Global} visual cues are obtained by max pooling over the whole feature layer.

\noindent \textbf{Grasp features.}
Our grasp classification network is trained on the GUN71 dataset~\cite{rogez2015understanding}. GUN71 contains $12,000$ RGB-D images, labeled with the corresponding grasp types. There are 71 grasp types in total.
We use our previously trained hand detector to estimate \emph{hand-context} regions, and crop and resize these regions to a size of $350 \times 350$. When there are two or more hand detections, for robustness we use only the highest scoring hand detection. We start with a model pre-trained on ImageNet~\cite{deng2009imagenet}, and train it for 100K iterations with Adam, setting the learning rate to $0.0001$. 

To prevent overfitting, we randomly crop $299 \times 299$ patches from the original image, $350 \times 350$, apply horizontal flips, and distort contrast, hue, saturation and brightness values of the image. To further gain robustness against overfitting, we use dropout with a ratio of $0.2$ and  $L2$ regularization with a weight of $0.0001$.

As a sanity check, we compare the grasp accuracy obtained by our model against the state of the art~\cite{rogez2015understanding} in~\tabref{tbl:gun71_grasp}. We follow the same experiment settings as~\cite{rogez2015understanding}. We compute accuracy as the percentage of correctly classified grasps. Our CNN-based approach outperforms the method of~\cite{rogez2015understanding}, that uses pre-trained VGG~\cite{simonyan2014very} features and an SVM classifier, by a large margin.

\noindent \textbf{Optical flow.} \label{para:of}
We train our flow estimation network on EPIC, using the 12-class training set described above.
We pre-compute optical flow by using the TV-L1 algorithm proposed in~\cite{zach2007duality}, with the same parameters used in~\cite{damen2018eccv}. At training time we feed 3 consecutive frames to the network, and minimize the cross-entropy loss for action classification.
We use the Inception-ResNet-v2 architecture~\cite{szegedy2017inception}, adapting the input and output layer sizes according to the flow image size and the number of classes, and use \emph{Mixed\_6a} layer features from this network for our temporal model.

\begin{table*}[!t]
	\centering{
	\begin{tabular}{
			>{\raggedright\arraybackslash}p{4cm}|
			>{\columncolor{Res2}}>{\centering\arraybackslash}p{1.5cm}|
			>{\columncolor{Res2}}>{\centering\arraybackslash}p{1.5cm}|
			>{\columncolor{Res2}}>{\centering\arraybackslash}p{1.5cm}|
			>{\columncolor{Res1}}>{\centering\arraybackslash}p{1.5cm}|
			>{\columncolor{Res1}}>{\centering\arraybackslash}p{1.5cm}|
			>{\columncolor{Res1}}>{\centering\arraybackslash}p{1.5cm}} 
		\multicolumn{1}{c}{} & 
		\multicolumn{3}{c}{\small {EPIC: Inter-class}} &  
		\multicolumn{3}{c}{\small {EGTEA: Inter-dataset}}\\
		\cmidrule{2-7}
		\multicolumn{1}{c}{}
		&\multicolumn{1}{c}{KNN-1}& \multicolumn{1}{c}{KNN-10}& \multicolumn{1}{c}{KNN-20}
		&\multicolumn{1}{c}{KNN-1}& \multicolumn{1}{c}{KNN-10}& \multicolumn{1}{c}{KNN-20}\\
	  \midrule
	  I3D \cite{carreira17cvpr}  & 12.8 & 18.2& 19.4 & 12.3 & 19.9& 25.3 \\
	  Two-Stream I3D \cite{carreira17cvpr} & 13.7 & 21.7 & 25.1 & 24.8 & 25.6& 34.7\\
	  TRN \cite{zhou2018temporal}& 13.2 & 22.7& 27.9 & 18.2 & 33.7& 42.5 \\
	  Two-Stream TRN \cite{zhou2018temporal} & 14.3 & 26.4 & 29.4 & 20.3 & 35.7& 44.3\\
		\midrule		
		 Coco-Global & 12.1 $\pm$ 0.8 & 17.5 $\pm$ 1.1 & 20.4 $\pm$ 0.7 & 24.6 $\pm$ 3.1  & 35.5 $\pm$ 2.9 & 40.1 $\pm$ 2.1  \\
     Coco-Global + Flow-Global  & 14.5 $\pm$ 1.0 & 21.3 $\pm$ 1.7 & 25.2 $\pm$ 1.2 & 25.6 $\pm$ 3.6  & 37.3 $\pm$ 3.0 & 42.5 $\pm$ 2.0 \\
     \midrule
     Hand  & 16.5 $\pm$ 1.0 & 26.2 $\pm$ 1.5 & 30.7 $\pm$ 1.6 & 26.5 $\pm$ 1.8  & 37.2 $\pm$ 2.0 & 41.9 $\pm$ 1.8 \\
     + Obj& 16.9 $\pm$ 1.3 & 27.8 $\pm$ 1.1 & 32.2 $\pm$ 1.1 & \textcolor{blue}{29.8} $\pm$ 1.4 & \textbf{42.8}  $\pm$ 1.7 & \textcolor{blue}{48.0}  $\pm$ 1.2\\
     + Flow& 17.0 $\pm$ 1.4 & 27.6 $\pm$ 0.9 & 32.4 $\pm$ 0.7 & 25.9 $\pm$ 1.3 & 36.4  $\pm$ 1.6& 41.6 $\pm$ 1.7\\
     + Traj& 15.2 $\pm$ 1.4 & 22.7 $\pm$ 1.4 & 27.2 $\pm$ 0.5 & 27.2 $\pm$ 1.1 & 37.6 $\pm$ 1.4 & 42.5 $\pm$ 1.0 \\
     + Grasp & 16.9 $\pm$ 0.7 & 28.5 $\pm$ 0.9 & \textbf{36.7} $\pm$ 0.7 & 24.2 $\pm$ 1.3 & 33.7 $\pm$ 1.4 & 38.4 $\pm$ 1.6 \\ 
     + Obj + Flow & 17.0 $\pm$ 0.9 & 27.6 $\pm$ 1.1 & 32.4 $\pm$ 1.3 & 26.9 $\pm$ 1.2 & 39.1 $\pm$ 1.5 & \textbf{49.8} $\pm$ 1.5\\
     + Obj + Flow + Grasp  & \textbf{19.9} $\pm$ 0.8 & \textcolor{blue}{29.5} $\pm$ 1.2& \textcolor{blue}{35.1} $\pm$ 1.2 & 29.1 $\pm$ 1.1  & \textcolor{blue}{41.4} $\pm$ 1.2 & 46.6 $\pm$ 1.0 \\ 
     + Obj + Flow + Grasp+Traj  & \textcolor{blue}{18.7} $\pm$ 1.0 & \textbf{30.3} $\pm$ 1.4& 34.6 $\pm$ 0.9 &\textbf{31.5} $\pm$ 1.0& 41.0 $\pm$ 1.2 & 47.0 $\pm$ 1.3 \\ 

		\bottomrule
	\end{tabular}}
	\setlength{\belowcaptionskip}{\RemoveBelowCaption}
	\vspace{4mm}
	\caption{
	Classification accuracy (in \%) for transfer learning experiments with K-Nearest Neighbors. During test time, we select respectively 1, 10, and 20 samples per class for KNN-1, KNN-10, KNN-20 experiments and then assign labels to test samples via majority voting of the labels of 1, 10, and 20 nearest samples. The best and second best results are highlighted with {\bf bold} and \textcolor{blue}{blue} fonts, respectively.} \label{tbl:knn}
		\vspace{4mm}
\end{table*}

\noindent \textbf{Temporal model.}
We train our temporal model on the same set of videos used for our flow network. We subdivide each video into segments, so that no segment is longer than 12 seconds, whereas for efficiency each mini-batch is constructed out of same-size sequences. We clip the norm of the gradients whenever they go over a threshold ($-25$ and $25$). We use cross entropy loss, 
\emph{L2} regularization and the attention penalty term $P$ (see Eq. \ref{eq:attn-norm}) with weights of $0.00005$ and $0.05$, respectively. \tabref{tbl:training_details} shows our temporal model training details. We use the same optimizer and learning rate also for fine tuning. In our model, training a single forward and backward pass takes around $45ms$ on an NVIDIA K-40 GPU.

\noindent \textbf{Baseline models for comparison with our approach.} We consider I3D~\cite{carreira17cvpr}, TRN~\cite{zhou2018temporal} and their two-stream versions as state-of-the-art baselines, using their publicly available implementation~\cite{carreira17cvpr,zhou2018temporal}. I3D and TRN take as input RGB image sequences; Two-Stream TRN and I3D take as input both RGB images and flow fields, thus requiring twice more parameters as I3D and TRN. As we did for our flow network training, we use the flow fields pre-computed with~\cite{zach2007duality}.
We resize the input images to 274 $\times$ 274; to prevent overfitting, we randomly crop them and modify their contrast, hue, saturation and brightness values. Both TRN~\cite{zhou2018temporal} and I3D~\cite{carreira17cvpr} are trained for 150K iterations. I3D and TRN, and their two-stream versions, require fixed input video length at training time; therefore, we subdivide each sequence into $15$ segments and randomly pick one segment per sequence.  More details about the training process can be seen in~\tabref{tbl:training_details}. At test time, we run the model 10 times and select an output action class via majority voting.

\noindent \textbf{Training details.} 
In \tabref{tbl:training_details}, we present full training details for hand detection, grasp classification, flow estimation, and our temporal model along with those for the baselines we consider, \ie~I3D~\cite{carreira17cvpr} and TRN~\cite{zhou2018temporal}. While comparing against~\cite{zhou2018temporal}, we use the multi-scale version of TRN~\cite{zhou2018temporal} as baseline, which reported  state-of-art accuracy in various datasets~\cite{zhou2018temporal}. As suggested by~\cite{zhou2018temporal}, we use all the TRN modules from 2-frame up to 8-frame. We use the default parameters for TRN with their publicly available code for training. The training code is not provided for~\cite{carreira17cvpr}, therefore we implemented our own. We set the training parameters such that the performance is optimized on the standard intra-dataset action recognition experiment on~\cite{damen2018eccv} and re-used the same training protocol for all the experiments. We initialized the model parameters from pre-trained models on the Kinetics dataset~\cite{carreira17cvpr}. All our models were implemented in TensorFlow~\cite{abadi2016tensorflow}.

\begin{table*}[!t]
	\centering{
	\begin{tabular}{
			>{\raggedright\arraybackslash}p{4cm}|
			>{\columncolor{Res1}}>{\centering\arraybackslash}p{1.5cm}|
			>{\columncolor{Res1}}>{\centering\arraybackslash}p{1.5cm}|
			>{\columncolor{Res1}}>{\centering\arraybackslash}p{1.5cm}|
			>{\columncolor{Res2}}>{\centering\arraybackslash}p{1.5cm}|
			>{\columncolor{Res2}}>{\centering\arraybackslash}p{1.5cm}|
			>{\columncolor{Res2}}>{\centering\arraybackslash}p{1.5cm}} 
		\multicolumn{1}{c}{} & 
		\multicolumn{3}{c}{\small {EPIC: Inter-class}} &  
		\multicolumn{3}{c}{\small {EGTEA: Inter-dataset}} \\
		\cmidrule{2-7}
		\multicolumn{1}{c}{\small {No. of Training Examples}}
		&\multicolumn{1}{c}{1} 
		&\multicolumn{1}{c}{10}
		&\multicolumn{1}{c}{30}
		&\multicolumn{1}{c}{1}
		&\multicolumn{1}{c}{10} 
		&\multicolumn{1}{c}{30}
		\\
	  \midrule
	  I3D \cite{carreira17cvpr} & 13.2 & 20.1 & 23.6 & 15.7& 34.4 & 36.3 \\
	  Two-Stream I3D \cite{carreira17cvpr} & 14.8 & 23.4 & 31.8 & 19.8 & 38.0 & 43.5 \\
	  TRN \cite{zhou2018temporal} & 16.1 & 27.5 & 35.7 & 23.5& 51.1 & 58.2 \\
	  Two-Stream TRN \cite{zhou2018temporal}  & 18.7 & 29.7 & \textcolor{blue}{39.3} & 26.2 & 55.8 & \textbf{62.5} \\
		\midrule
	Coco-Global & 10.9 $\pm$ 1.4 & 16.9 $\pm$ 0.7 & 26.5 $\pm$ 1.2 & 15.9 $\pm$ 2.9  & 43.3 $\pm$ 1.1  & 51.0 $\pm$ 0.6  \\
     Coco-Global + Flow-Global & 13.8 $\pm$ 1.2  & 22.1 $\pm$ 0.8  & 30.0 $\pm$ 1.0  & 21.0 $\pm$ 2.3  & 46.2 $\pm$ 1.1  & 52.3 $\pm$ 0.5  \\
     \midrule
     Hand & 16.4 $\pm$ 1.2  & 27.6 $\pm$ 0.7  & 34.8 $\pm$ 0.9  & 21.7 $\pm$ 2.7  & 48.4 $\pm$ 1.5  & 56.9 $\pm$ 0.6  \\
     + Obj & 18.1 $\pm$ 1.3  & 28.3 $\pm$ 0.7   & 36.4 $\pm$ 0.8  & 29.7 $\pm$ 2.1  & 50.6 $\pm$ 1.3 & 57.7 $\pm$ 0.3  \\
     + Flow & 17.9 $\pm$ 1.6  & 26.9 $\pm$ 0.8  & 35.6 $\pm$ 0.6  & 22.7 $\pm$ 2.8  & 53.8 $\pm$ 1.2  & 58.2 $\pm$ 0.2  \\
     + Traj & 15.2 $\pm$ 1.3  & 27.1 $\pm$ 0.8  & 35.0 $\pm$ 0.7  & 21.8 $\pm$ 2.6  & 51.4 $\pm$ 1.4  & 57.3  $\pm$ 0.7 \\
     + Grasp & 19.7 $\pm$ 1.4  & \textcolor{blue}{29.8} $\pm$ 0.4  & 38.1 $\pm$ 0.4  & 23.8 $\pm$ 2.7  & 52.7 $\pm$ 1.9  & 58.5 $\pm$ 0.5  \\ 
     + Obj + Flow & 17.8 $\pm$ 1.2  & 27.0 $\pm$ 0.5  & 37.9 $\pm$ 0.7 & 28.3 $\pm$ 2.3  & 54.5 $\pm$ 1.3  & 59.0 $\pm$ 0.1 \\
     + Obj + Flow + Grasp & \textbf{20.6} $\pm$ 1.3  & 29.7 $\pm$ 0.4  & \textbf{39.8} $\pm$ 0.6 & \textcolor{blue}{30.9} $\pm$ 2.0  & \textcolor{blue}{56.9} $\pm$ 1.3  & 60.4 $\pm$ 0.2 \\
        + Obj + Flow + Grasp+Traj & \textcolor{blue}{20.3} $\pm$ 1.2  & \textbf{31.7} $\pm$ 0.4  & \textcolor{blue}{38.5} $\pm$ 0.4 & \textbf{32.0} $\pm$ 2.3  & \textbf{57.1} $\pm$ 1.4  & \textcolor{blue}{61.2} $\pm$ 0.3 \\

		\bottomrule
	\end{tabular}}
	\setlength{\belowcaptionskip}{\RemoveBelowCaption}
	\vspace{4mm}
	\caption{
Classification accuracy (in \%) for transfer learning experiments: here we fine tune all the network parameters of our temporal model (RNN + attention layer + final logits) and baselines. The best and second best results are highlighted with {\bf bold} and \textcolor{blue}{blue} fonts, respectively.		\label{tbl:fine-tune}} 
\vspace{5mm}
\end{table*}

\begin{table*}[!t]
	\centering{
	\begin{tabular}{
			>{\raggedright\arraybackslash}p{4cm}|
			>{\columncolor{Res1}}>{\centering\arraybackslash}p{1.5cm}|
			>{\columncolor{Res1}}>{\centering\arraybackslash}p{1.5cm}|
			>{\columncolor{Res1}}>{\centering\arraybackslash}p{1.5cm}|
			>{\columncolor{Res2}}>{\centering\arraybackslash}p{1.5cm}|
			>{\columncolor{Res2}}>{\centering\arraybackslash}p{1.5cm}|
			>{\columncolor{Res2}}>{\centering\arraybackslash}p{1.5cm}} 
		\multicolumn{1}{c}{} & 
		\multicolumn{3}{c}{\small {EPIC: Inter-class}} &  
		\multicolumn{3}{c}{\small {EGTEA: Inter-dataset}} \\
		\cmidrule{2-7}
		\multicolumn{1}{c}{\small {No. of Training Examples}}
		&\multicolumn{1}{c}{1} 
		&\multicolumn{1}{c}{10}
		&\multicolumn{1}{c}{30}
		&\multicolumn{1}{c}{1}
		&\multicolumn{1}{c}{10} 
		&\multicolumn{1}{c}{30}
		\\
	  \midrule
	  I3D \cite{carreira17cvpr} & 13.8 & 19.7 & 24.3 & 14.5 & 29.7 & 31.1 \\
	  Two-Stream I3D \cite{carreira17cvpr} & 14.6 & 21.1 & 29.9 & 18.1 & 34.3 & 40.2 \\
	 TRN \cite{zhou2018temporal} & 15.2 & 24.3 & 31.6 & 16.9 & 32.1 & 35.7 \\
	  Two-Stream TRN \cite{zhou2018temporal} & 17.1 & 25.8 & 34.5 & 19.3 & 35.9 & 45.4 \\
		\midrule
	Coco-Global & 11.9 $\pm$ 1.2 & 16.5 $\pm$ 0.2 & 21.7 $\pm$ 1.2 & 17.2 $\pm$ 2.9  & 40.2 $\pm$ 1.1  & 47.5 $\pm$ 1.2  \\
     Coco-Global + Flow-Global & 13.4 $\pm$ 1.1  & 19.1 $\pm$ 0.3  & 26.7 $\pm$ 1.0  & 23.6 $\pm$ 3.0  & 43.9 $\pm$ 1.1  & 50.2 $\pm$ 1.1  \\
     \midrule
     Hand & 16.7 $\pm$ 1.3  & 27.2 $\pm$ 0.3  & 31.1 $\pm$ 0.3  & 26.0 $\pm$ 3.7  & 44.8 $\pm$ 0.9  & 50.9 $\pm$ 1.2  \\
     + Obj & 17.2 $\pm$ 1.4  & 27.8 $\pm$ 0.2   & 34.0 $\pm$ 1.0  & 26.2 $\pm$ 2.7  & 48.5 $\pm$ 0.4 & 56.3 $\pm$ 0.9  \\
     + Flow & 16.8 $\pm$ 1.3  & 25.4 $\pm$ 0.2  & 31.5 $\pm$ 0.4  & 22.5 $\pm$ 2.8  & 45.5 $\pm$ 0.6  & 54.4 $\pm$ 1.3  \\
     + Traj & 17.5 $\pm$ 1.3  & 25.7 $\pm$ 0.2  & 32.9 $\pm$ 0.2  & 24.1 $\pm$ 1.9  & 46.7 $\pm$ 0.7  & 52.1  $\pm$ 1.2 \\
     + Grasp & 18.5 $\pm$ 1.3  & 28.9 $\pm$ 0.1  & \textcolor{blue}{36.0} $\pm$ 0.1  & 20.0 $\pm$ 1.7  & 45.2 $\pm$ 1.2  & 53.2 $\pm$ 0.7  \\ 
     + Obj + Flow & 16.3 $\pm$ 1.2  & 25.6 $\pm$ 0.3  & 35.9 $\pm$ 0.9 & 24.5 $\pm$ 1.9  & \textbf{50.1} $\pm$ 1.3  & 55.4 $\pm$ 0.9 \\
     + Obj + Flow + Grasp & \textcolor{blue}{19.8} $\pm$ 1.4  & \textcolor{blue}{29.0} $\pm$ 0.2  & 35.2 $\pm$ 0.6 & \textbf{30.0} $\pm$ 2.0  & \textcolor{blue}{49.2} $\pm$ 1.2  & \textbf{56.7} $\pm$ 0.7 \\
     + Obj + Flow + Grasp+Traj & \textbf{20.0} $\pm$ 1.5  & \textbf{31.5} $\pm$ 0.2  & \textbf{36.9} $\pm$ 0.8 & \textcolor{blue}{29.5} $\pm$ 2.3  & 48.9 $\pm$ 1.5  & \textcolor{blue}{55.8} $\pm$ 0.8 \\

		\bottomrule
	\end{tabular}}
	\setlength{\belowcaptionskip}{\RemoveBelowCaption}
	\vspace{4mm}
	\caption{Classification accuracy (in \%) for transfer learning experiments for which we fine tune only the final logit layers.}
		\vspace{4mm}
		\label{tbl:fine-tune-logits}
\end{table*}

\begin{figure*}[t]
	\centering
	\includegraphics[width=1\linewidth]{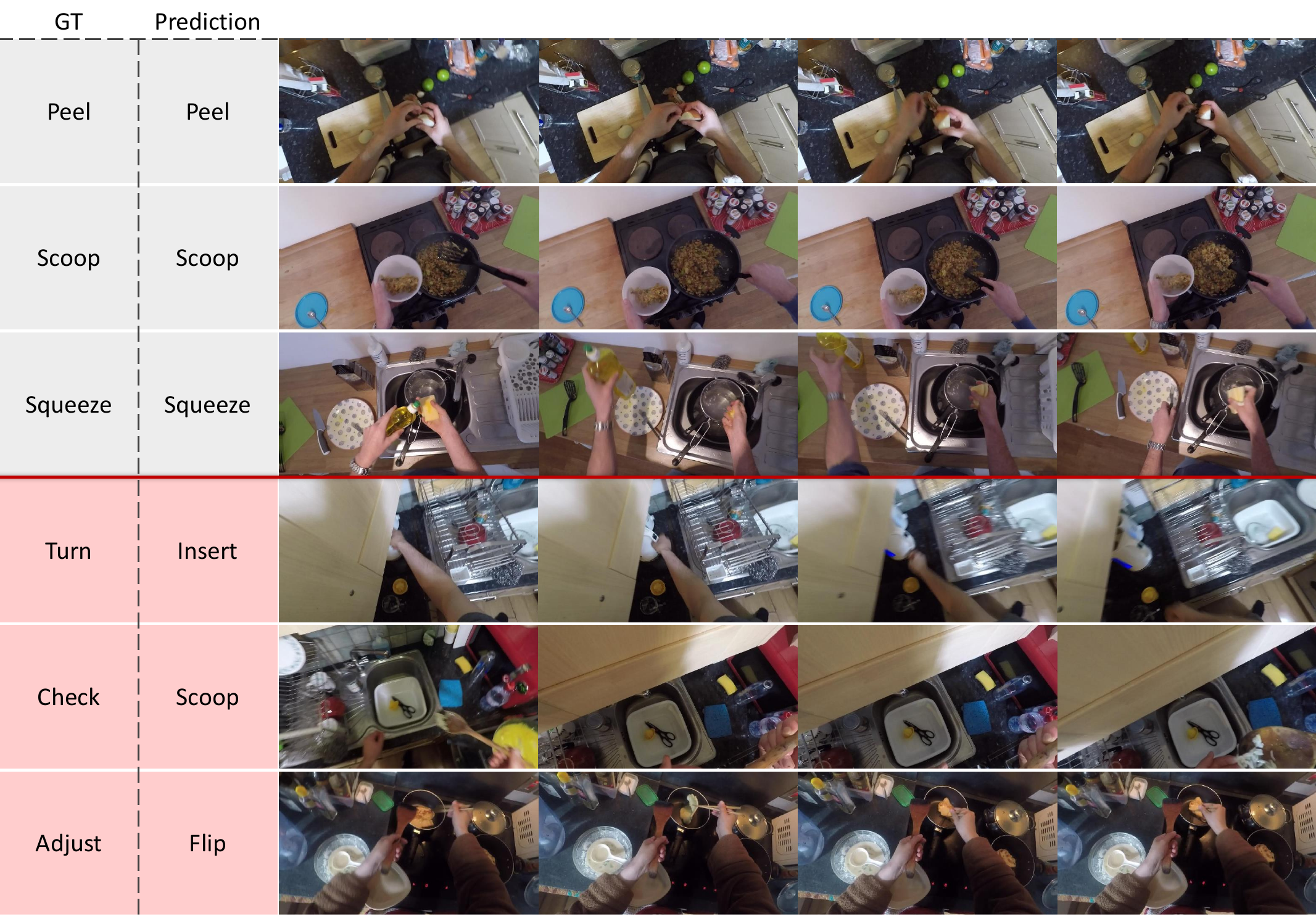}
	\caption{Qualitative results for our inter-class experiments on EPIC. Frames are selected uniformly at random; different rows correspond to different activities. The first three rows show correct predictions; the last three rows highlight failure cases.
	Even when wrong, our predictions still provide a reasonable explanation of the scene. 
	}	
	\label{fig:qual}
\end{figure*}

\subsection{Inter-dataset transfer} \label{sec:inter-dataset} 
Inter-dataset transfer uses an action classification model trained on one dataset and tests it with a few examples on an entirely new dataset.
We use the EGTEA dataset \cite{li13} for our inter-dataset transfer experiments. We subsampled the original sequences from 24 to 12 fps, and use the train and test splits provided by the dataset. Our training set consists of 12 classes from EPIC (\emph{close}, \emph{cut}, \emph{mix}, \emph{move}, \emph{open}, \emph{pour}, \emph{put}, \emph{remove}, \emph{take}, \emph{throw}, \emph{turn-on}, \emph{wash}); our test set consists of 10 classes from EGTEA (\ie, \emph{close}, \emph{cut}, \emph{mix}, \emph{move}, \emph{open}, \emph{pour}, \emph{put}, \emph{take}, \emph{turn-on}, \emph{wash}). 
Optical flow is computed via TV-L1 \cite{zach2007duality}. \tabref{tbl:knn},  \tabref{tbl:fine-tune} and \tabref{tbl:fine-tune-logits} report the results obtained by using KNN, fine tuning all the parameters and fine tuning only the logits parameters, respectively.  \tabref{tbl:knn} reports the percentage of correctly classified actions using KNN with K equal to 1, 10, and 20.  \tabref{tbl:fine-tune} and  \tabref{tbl:fine-tune-logits} show the percentage of correct classifications in the \emph{fine tuning} experiment: in this case, each classification is obtained from the network's softmax, again tested in the three cases of 1, 10, 30 selected samples from each class.

We obtain the best results by combining \emph{Hand}, \emph{Obj}, \emph{Flow}, \emph{Grasp} and  \emph{Traj} features for 1 and 10 samples. For the case of 30 samples, Two-Stream TRN obtains the best results. This demonstrates the superior performance with the effective segmentation of the activity from the background carried out by the employed visual cues especially for few shot learning scenarios. 

Intuitively, we can argue that object-related cues (\ie, \emph{Obj}) provide the biggest contribution in terms of performance since similar activities deploy similar objects. 
Also, we can observe that using trajectory-related cues consistently improves accuracy in all cases.
The poor performance of flow-related cues in the KNN experiment can be explained by the fact that the two datasets significantly differ in terms of recording settings and camera setup: EPIC is recorded with a head-mounted GoPro, while EGTEA is recorded using eye-tracking glasses. Note that, in the \emph{fine tuning} experiment, the \emph{Flow} cue is the second best cue in terms of accuracy after the \emph{Grasp} cue.

Comparing the accuracy for methods that fine tune all the network parameters (\tabref{tbl:fine-tune}) and only the logit layers (\tabref{tbl:fine-tune-logits}), we see that the former yields higher accuracy in almost all cases.

\subsection{Inter-class Transfer} \label{sec:inter-class}

Inter-class transfer uses frequent classes in a dataset to train a model which is then tested on classes with sparse data. We conduct an \emph{inter-class} transfer experiment on EPIC by considering 14 classes (namely \emph{adjust}, \emph{check}, \emph{dry}, \emph{empty}, \emph{fill}, \emph{flip}, \emph{insert}, \emph{peel}, \emph{press}, \emph{scoop}, \emph{shake}, \emph{squeeze}, \emph{turn}, \emph{turn-off}) . Note that our training set consists of 12 disjoint classes from EPIC (\emph{close}, \emph{cut}, \emph{mix}, \emph{move}, \emph{open}, \emph{pour}, \emph{put}, \emph{remove}, \emph{take}, \emph{throw}, \emph{turn-on}, \emph{wash}). The results are shown in \tabref{tbl:knn} and \tabref{tbl:fine-tune}. 
We follow the same experiment protocol described in \secref{sec:inter-dataset}.

Our KNN method clearly outperforms the state of the art by a large margin. In particular, the \emph{Hand + Grasp} combination yields a 36.7\% accuracy using 20 nearest samples: this is more than a 7.0 percentage improvement on the performance of the Two-Stream TRN, that scores 29.4\%. A similar trend is exhibited in the \emph{fine tuning} experiment, in this case the best combination is \emph{Hand + Obj + Flow + Grasp} (\ie, 39.8\%), showing the importance of combining together multiple heterogeneous features. 

We observe that grasp contributes the most to \emph{inter-class} transfer accuracy, as the grasp type used for a specific object is quite discriminative for the specific activity being carried out, so that the temporal model can learn to distinguish actions by looking only at a small number of training samples. 
Also, global features seem to under-perform in this case: this also highlights the importance of using the hand as a hard attention mechanism.

\subsection{Intra-dataset Transfer} \label{sec:intra-data}
We conduct an \emph{intra-dataset} experiment on EPIC following the same experiment protocol described in \secref{sec:inter-dataset}.  \tabref{tbl:knn-learned-classes} reports results for our visual cues when training and test sets have the same activity labels (namely, \emph{close}, \emph{cut}, \emph{mix}, \emph{move}, \emph{open}, \emph{pour}, \emph{put}, \emph{remove}, \emph{take}, \emph{throw}, \emph{turn-on}, \emph{wash}), from EPIC. We  compare against I3D~\cite{carreira17cvpr}, Two Stream I3D~\cite{carreira17cvpr}, TRN~\cite{zhou2018temporal} and Two Stream TRN~\cite{zhou2018temporal}, trained as described above. In addition, we design two further baseline methods: \emph{Coco-Global} and \emph{Coco-Global+Flow-Global}. Here, \emph{Coco-Global} extracts global features from the input RGB frame (obtained from an Inception-ResNet-v2 network trained for MS-COCO~\cite{coco} detection, \emph{Mixed\_6a} features), whereas \emph{Coco-Global+Flow-Global} extracts additional global features from optical flow, plugging these features into our temporal model. These baselines mimic most state-of-the-art approaches to action recognition, which feed similar global features into their respective temporal models~\cite{singh2016cvpr,lea2017cvpr}. 
As we can see in \tabref{tbl:knn-learned-classes}, our complete pipeline outperforms the compared approaches by significant margins, \emph{i.e.} $7.5$ percentage points (pp) and $1.6$ pp against the best baseline model for KNN and Softmax classifiers, respectively. We observe that fine-tuning leads to very high standard deviations when we use only 1 sample per class (L=1); this can be explained by the model overfitting to data and getting stuck in local minima.

\begin{table*}[!t]
	\centering{
	\begin{tabular}{
			>{\raggedright\arraybackslash}p{4.2cm}|
			>{\columncolor{Res1}}>{\centering\arraybackslash}p{1.5cm}|
			>{\columncolor{Res1}}>{\centering\arraybackslash}p{1.5cm}|
			>{\columncolor{Res1}}>{\centering\arraybackslash}p{1.5cm}|
			>{\columncolor{Res2}}>{\centering\arraybackslash}p{1.5cm}} 
		\multicolumn{1}{c}{} & 
		\multicolumn{4}{c}{\small {EPIC: Intra-dataset}} \\
		\cmidrule{2-5}
		\multicolumn{1}{c}{}
		& \multicolumn{1}{c}{KNN-1}
		& \multicolumn{1}{c}{KNN-10}
		& \multicolumn{1}{c}{KNN-20} 
		& \multicolumn{1}{c}{SftMx} \\
	  \midrule
	  I3D \cite{carreira17cvpr} & 16.3 & 20.9 & 26.0 & 42.3\\
	  Two-Stream I3D \cite{carreira17cvpr} & 17.5 & 27.9& 32.11 & 54.2 \\
	  TRN \cite{zhou2018temporal} & 18.0 & 30.2 & 34.0 & 54.5 \\
	  Two-Stream TRN \cite{zhou2018temporal} & 21.4 & 34.0 & 40.3 & 60.6\\
		\midrule		
	 Coco-Global & 21.2 $\pm$ 2.1 & 34.5 $\pm$ 1.8 & 40.0 $\pm$ 1.9 & 41.7 $\pm$ 1.1  \\
     Coco-Global + Flow-Global & 25.4 $\pm$ 1.9 & 37.7 $\pm$ 1.6 & 48.9 $\pm$ 1.4 & 52.2 $\pm$ 1.0 \\
     \midrule
     Hand & 26.1 $\pm$ 1.5 & 37.6$\pm$ 1.4  & 48.1 $\pm$ 1.0 & 52.2 $\pm$ 0.5 \\
     + Obj & 28.7 $\pm$ 1.9 & 39.6$\pm$ 1.8 & 50.0 $\pm$ 1.1 & 55.9 $\pm$ 0.9 \\
     + Flow & 29.6 $\pm$ 1.8 & 41.1$\pm$ 1.3  & 53.2 $\pm$ 0.8  & 57.5 $\pm$ 0.4\\
     + Traj & 29.8 $\pm$ 1.4 & 42.0$\pm$ 1.1  & 49.7 $\pm$ 1.2 & 53.7 $\pm$ 0.6 \\
     + Grasp & 28.3 $\pm$ 0.9 & 42.5$\pm$ 0.8  & 49.3 $\pm$ 0.5 & 54.8  $\pm$ 0.2\\
     + Obj + Flow & 30.0 $\pm$ 1.1 & \textcolor{blue}{46.9}$\pm$ 1.0  & 54.7  $\pm$ 0.5 & 59.9 $\pm$ 0.1\\
     + Obj + Flow + Grasp & \textcolor{blue}{31.8} $\pm$ 1.6 & \textcolor{blue}{46.9} $\pm$ 0.7  & \textcolor{blue}{56.4} $\pm$ 0.8 &  \textcolor{blue}{61.6} $\pm$ 1.0\\ 
     + Obj + Flow + Grasp + Traj & \textbf{32.2} $\pm$ 1.4 & \textbf{48.1} $\pm$ 0.7 & \textbf{56.9} $\pm$ 0.5 & \textbf{62.2} $\pm$ 0.9\\ 
		\bottomrule
	\end{tabular}}
	\setlength{\belowcaptionskip}{\RemoveBelowCaption}
	\vspace{4mm}
	\caption{Classification accuracy (in \%) on the EPIC dataset test subjects. In this task we trained and tested on the same classes. KNN-1: 1 nearest neighbor, KNN-10: 10 nearest neighbor, KNN-20: 20 nearest neighbor, SftMx: Softmax.
		\label{tbl:knn-learned-classes}}
\end{table*}

\subsection{Learning to Transfer}
\label{sec:meta-learning}
\begin{figure}[t]
	\centering
	\scalebox{1}{
	\includegraphics[clip, trim=0cm 0cm 0cm 0cm, width=.97\linewidth]{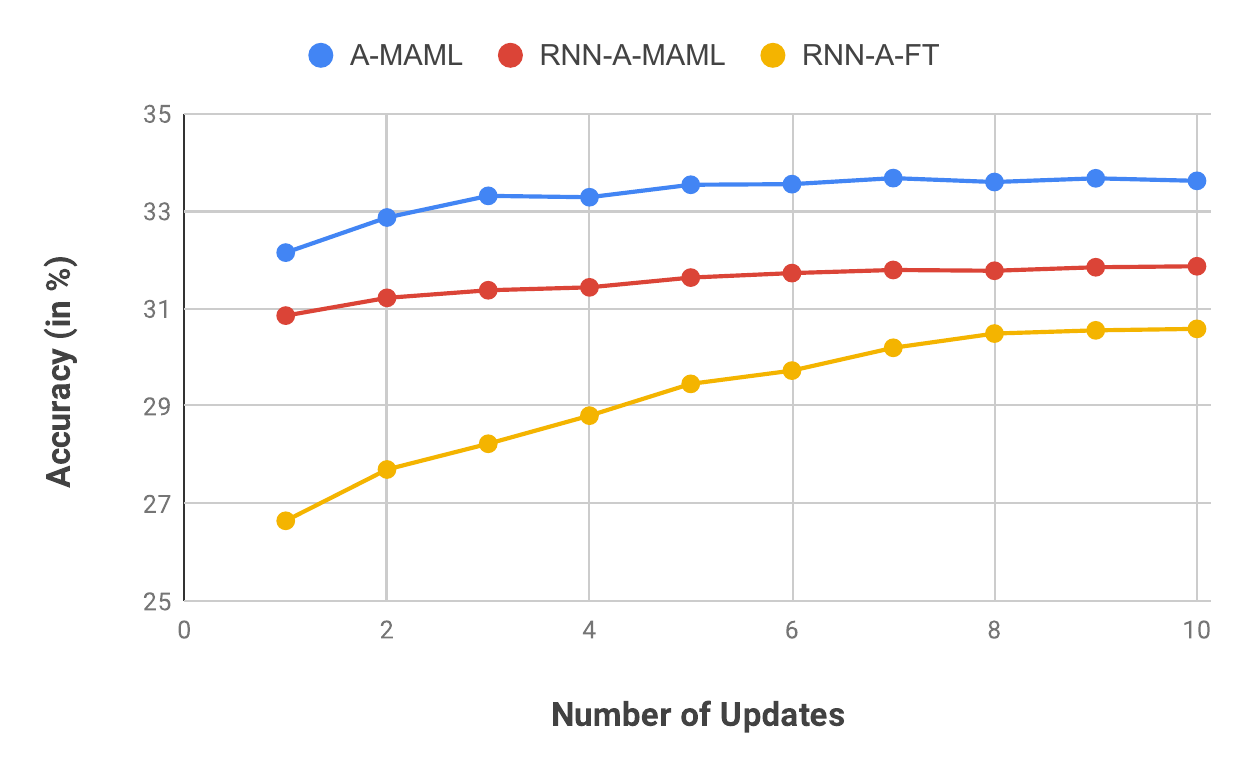}}
	\caption{Convergence of models at test-time as a function of the number of updates of the model parameters. All training strategies use the full set of visual cues and identical models.}
	\label{fig:nupdates}
\end{figure}

We further assess the effectiveness of different transfer learning strategies and demonstrate that our attentive meta transfer-learning model, A-MAML, achieves better accuracy than competing transfer learning strategies for few-shot action recognition.~\tabref{tbl:meta_learng} demonstrates our results for the 1-, 5-, and 10-shot 5-class action recognition experiment on the EPIC dataset. We use 12 classes from EPIC (namely, \emph{close}, \emph{cut}, \emph{mix}, \emph{move}, \emph{open}, \emph{pour}, \emph{put}, \emph{remove}, \emph{take}, \emph{throw}, \emph{turn-on}, \emph{wash}) for training and 14 classes (namely \emph{adjust}, \emph{check}, \emph{dry}, \emph{empty}, \emph{fill}, \emph{flip}, \emph{insert}, \emph{peel}, \emph{press}, \emph{scoop}, \emph{shake}, \emph{squeeze}, \emph{turn}, \emph{turn-off}) for testing.
During training meta-learning models, $5$ classes are randomly selected from which $5$ samples are drawn for the learner updates. Hence, we evaluate our few-shot learning accuracy for the 5-way, 5-shot classification setting. The meta batch-size is set to $10$ tasks. While the learner is trained with 5 steps, the meta-learner is trained for 2K steps. We use vanilla gradient descent with a learning rate of $0.001$ within a specific task and Adam with a learning rate of $0.001$ for the meta-learner updates. We evaluate all methods on 100 different tasks. Each task consists of 5 randomly selected distinct classes, drawn from the test classes, and each class contains k (k=1,5,10) samples. We save the 100 generated tasks and use this for all the models, in order to ensure a fair comparison between models. \tabref{tbl:meta_learng} shows that meta-learning based transfer learning outperforms the baselines by a large margin and the proposed A-MAML training further improves the accuracy on all feature combinations in our experiments. While training attention jointly with RNN (RNN-A-MAML) yields compelling action recognition accuracy, the simpler model that relies on only attention-based training yields the best accuracy (A-MAML). Ultimately, our attention-based model provides a more structured memory for handling long-term dependencies compared to RNNs~\cite{radford2018improving}, and yields robust transfer performance across diverse tasks. We can observe that the classification accuracy significantly drops for fine-tuning based approaches as compared to meta-learning based approaches, when the number of training samples is low. In fact, for the case of one-shot action action classification, our attentive meta-learning strategy yields the best accuracy. The performance improvement of A-MAML with respect to baselines is  more pronounced in the case of 1-shot 5-way classification as compared to 5-shot 5-way and 10-shot 5-way classification. This suggests that A-MAML is robust even when the number of samples is fewer (\emph{e.g.} 1 sample) per class, being less prone to overfitting. As we can see in  \figref{fig:nupdates}, A-MAML yields the best accuracy as compared to the baselines for any number of updates at test time. Furthermore, A-MAML achieves substantially higher accuracy than the baseline fine-tuning approach only with a single update.

\begin{table*}[t]
	\centering \scalebox{0.9} {
	\begin{tabular}{
			>{\raggedright\arraybackslash}p{0.9cm}|
			>{\raggedright\arraybackslash}p{4.2cm}|
			>{\columncolor{Res1}}>{\centering\arraybackslash}p{1.9cm}|
			>{\columncolor{Res2}}>{\centering\arraybackslash}p{1.7cm}|
			>{\columncolor{Res1}}>{\centering\arraybackslash}p{1.6cm}|
			>{\columncolor{Res2}}>{\centering\arraybackslash}p{2.2cm}|
			>{\columncolor{Res1}}>{\centering\arraybackslash}p{2.2cm}|
			>{\columncolor{Res2}}>{\centering\arraybackslash}p{2.6cm}|
			>{\columncolor{Res1}}>{\centering\arraybackslash}p{1.6cm}} 
		\multicolumn{1}{c}{nShot}
	  	&\multicolumn{1}{c}{Feature Set}
		&\multicolumn{1}{c}{RNN-A-KNN} 
		&\multicolumn{1}{c}{RNN-FT}
		&\multicolumn{1}{c}{RNN-A-FT}
		&\multicolumn{1}{c}{RNN-MAML}
		&\multicolumn{1}{c}{RNN-A-MAML}
		&\multicolumn{1}{c}{A-MAML} 
		\\
	  \midrule

       & Coco-Global + Flow-Global & 27.7 & 25.4 
	  & \textcolor{blue}{26.9} 
	  & 24.9  
	  &  26.3 
	  &  \textbf{28.5}   \\
    1-Shot & Ours (Hand Only)  & 26.1 & 26.3 
      & 26.5 
      & 28.4   
      & \textcolor{blue}{29.2}
      &  \textbf{30.9}  \\
     & Ours (All)  & 30.3 & 28.8 
      & 30.7 
      & 27.1   
      &  \textcolor{blue}{31.7} 
      &  \textbf{33.5}  \\
      \midrule
      
      & Coco-Global + Flow-Global & 29.9 & 30.2 
	  & 31.3 
	  & 33.5  
	  &  \textcolor{blue}{35.2} 
	  &  \textbf{35.7}   \\
    5-Shot & Ours (Hand Only)  & 30.8 & 31.1 
      & 32.8 
      & 34.6   
      & \textcolor{blue}{35.3}
      &  \textbf{36.9}  \\
    & Ours (All)  & 34.5 & 35.0 
      & 36.1 
      & 38.6   
      &  \textcolor{blue}{40.2} 
      &  \textbf{41.4}  \\
      
      \midrule
      & Coco-Global + Flow-Global & 32.9 & 39.2 
	  & 40.1 
	  & 41.2  
	  &  \textcolor{blue}{43.4} 
	  &  \textbf{44.5}   \\
    10-Shot & Ours (Hand Only)  & 33.6 & 40.8 
      & \textbf{44.8}
      & 40.7   
      & 41.8
      &  \textcolor{blue}{44.0}  \\
     & Ours (All)  & 39.3 & 46.1
      & \textcolor{blue}{49.7} 
      & 46.8   
      & 49.2
      &  \textbf{50.2}  \\
 	\bottomrule
	\end{tabular}}
	\setlength{\belowcaptionskip}{\RemoveBelowCaption}
	\vspace{4mm}
	\caption{Classification accuracy (in \%) for comparison of different transfer learning strategies on 1-shot 5-class,  5-shot 5-class, and 10-shot 5-class action recognition experiments on EPIC. Our attention-based meta learning approach, A-MAML, significantly outperforms other transfer learning strategies. For \emph{KNN} and \emph{FT}, we follow the standard training procedure described in Sec.~\ref{sec:train_transf}. Methods in columns [1,3,5,6] use identical models. For columns [2,4], the methods use models without attention layer. The best and second best results are highlighted with {\bf bold} and \textcolor{blue}{blue} fonts, respectively.
	{\em Notations: A: Attention, FT: Fine Tuning}.
		\label{tbl:meta_learng}} 
\end{table*}

In \figref{fig:qual} we visualize examples of our model predictions for 4 different activities for both correct predictions and failure cases. We note that some annotations in the dataset are confusing or wrongly annotated, for instance the third row depicts an example of the \emph{turn} activity (the subject is turning on the kettle), while such activity is defined for most examples in the dataset as that of physically turning or rotating an object.
This confusion is also label-wise other than sample-wise: \eg, in addition to the \emph{turn} label, the dataset also contains the \emph{turn-on} label.

\begin{table*}[!t]
	\centering{
	\begin{tabular}{
			>{\raggedright\arraybackslash}p{5cm}|
			>{\columncolor{Res1}}>{\centering\arraybackslash}p{1.1cm}|
			>{\columncolor{Res1}}>{\centering\arraybackslash}p{1.1cm}|
			>{\columncolor{Res1}}>{\centering\arraybackslash}p{1.1cm}|
			>{\columncolor{Res2}}>{\centering\arraybackslash}p{1.1cm}|
			>{\columncolor{Res2}}>{\centering\arraybackslash}p{1.1cm}|
			>{\columncolor{Res2}}>{\centering\arraybackslash}p{1.1cm}|
			>{\columncolor{Res2}}>{\centering\arraybackslash}p{1.1cm}}
		
	   	\multicolumn{1}{c}{} & 
		\multicolumn{3}{c}{\small {EPIC}} &  
		\multicolumn{3}{c}{\small {EGTEA}}\\
		\cmidrule{2-7}
		\multicolumn{1}{c}{}
		&\multicolumn{1}{c}{1-shot} 
		&\multicolumn{1}{c}{5-shot} 
		&\multicolumn{1}{c}{10-shot} 
		&\multicolumn{1}{c}{1-shot} 
		&\multicolumn{1}{c}{5-shot} 
		&\multicolumn{1}{c}{10-shot} 
		\\
	  \midrule
	   ProtoNet \cite{snell2017prototypical}  & 28.8 & 39.0 & 47.9   & 28.4 & 44.7 & 57.2\\
	   Baseline++ \cite{chen2019closer}  & 31.4 & \textcolor{blue}{41.2} & \textcolor{blue}{50.4}   & 30.4 & 49.9 & 60.3\\
	   
	   ProtoGAN \cite{kumar2019protogan}  & \textcolor{blue}{32.1} & 36.3 & 49.3   & \textbf{34.1} & 47.8 & 60.1 \\
	   TARN \cite{BishayZP19}  & 29.8 & 40.6 & \textbf{50.9}   & 27.5 & 48.2 & \textbf{62.0}\\
	   
     \midrule
       RNN-A-MAML~\cite{finn2017icml}    & 31.7   & 40.2  &49.4   & 31.6   & \textcolor{blue}{50.7}  & 59.9  \\
  
      A-MAML  & \textbf{32.9}   & \textbf{41.4}  & 50.2   & \textcolor{blue}{33.5}   & \textbf{51.4}  & \textcolor{blue}{60.7}  \\
		\bottomrule
	\end{tabular}}
	\setlength{\belowcaptionskip}{\RemoveBelowCaption}
	\vspace{4mm}
	\caption{
Classification accuracy (in \%) for 5 way classification task on EPIC and EGTEA datasets. The RNN-A-MAML model uses the same architecture as A-MAML, but it uses the naive MAML algorithm for the meta-update.}
		\vspace{4mm}
		\label{tbl:few-amaml-new}
\end{table*}

\textbf{Comparing with SOTA.}
We compare our A-MAML against several other methods on EPIC and EGTEA. For the experiments on EPIC, we use the same settings described above in this section.  For the EGTEA experiments, we use 10 classes (namely, \emph{close}, \emph{cut}, \emph{mix}, \emph{move-around}, \emph{open}, \emph{pour}, \emph{put}, \emph{take}, \emph{turn-on}, \emph{wash}) for training and 9 classes (namely, \emph{inspect-read}, \emph{operate}, \emph{spread}, \emph{turn-off}, \emph{divide-pull-apart}, \emph{clean-wipe}, \emph{compress}, \emph{crack}, \emph{squeeze}) for testing. Our 5-way, 1-shot,  5-shot, and 10-shot classification results can be seen in Table \ref{tbl:few-amaml-new}. 
In this experimental setup, our first baseline is the RNN-A-MAML~\cite{finn2017icml} approach which uses all our visual cues. Our second baseline is ProtoGan~\cite{kumar2019protogan}, which has reported SOTA results on various datasets. We train ProtoGan by closely following the training strategy described in the paper. Since this model requires intermediate video features, we extract them for the EPIC and EGTEA datasets by using a C3D network pre-trained on Sports-1M \cite{karpathy2014large}.
Our third baseline is TARN~\cite{BishayZP19}, which uses a deep metric learning  approach. We train TARN by following the settings recommended in the paper. Similarly to ProtoGan, TARN requires intermediate video features. We again use a C3D network pre-trained on Sports-1M~\cite{karpathy2014large}.
Our fourth baseline is Baseline++~\cite{chen2019closer}, which proposes a technique similar to fine-tuning and a loss function  designed to reduce intra-class variations. Our final baseline is ProtoNet~\cite{snell2017prototypical}. ProtoNet, similarly to TARN~\cite{BishayZP19}, uses a deep metric learning approach. It also aims to reduce the L2 distance between query samples and the support set in embedding space. Differently from TARN, support sets are constructed by computing the mean of the features that belong to the same class within a batch. To make a fair comparison, we use the same features as TARN~\cite{BishayZP19} and Protogan~\cite{kumar2019protogan} and use the official GitHub repositories for the implementation of the proposed loss functions. It can be seen from Table \ref{tbl:few-amaml-new} that A-MAML significantly outperforms all the baselines for 5-way and 5-shot classification tasks on both datasets. Notably, our method achieves compelling accuracy for few-shot classification tasks and attains the best accuracy for 1-shot and 5-shot classification on EPIC and 5-shot classification on EGTEA. It obtains the second best accuracy for 10-shot action classification. TARN and ProtoNet classify samples using a KNN matching approach, thus when the number of labeled samples per class is too small it fails to generalize well to novel classes, as can be seen from their results in Table~\ref{tbl:few-amaml-new} on 1- and 5-shot classification tasks. On the other hand, ProtoGan proposes a novel data augmentation strategy that yields good accuracy even when the number of samples is quite low. However, when the number of samples are high, the comparative advantage of ProtoGan is not as pronounced. Although it lags behind our method, Baseline++ performs reasonably well on both datasets, interestingly enough, considering the simplicity of the model. Note that the idea of reducing intra-class variations in Baseline++ is orthogonal to the framework we propose and can be combined with our approach.

\textbf{KNN vs Fine tuning} 
We observe that both KNN and fine-tuning perform comparably on the inter-class transfer task, \emph{e.g.} $29.5\%$ vs. $29.7\%$ for 10 samples in last rows of Tables~\ref{tbl:knn} and \ref{tbl:fine-tune} respectively. However, KNN is consistently outperformed by fine-tuning with a large margin on inter-dataset transfer, \emph{e.g.} $41.4\%$ vs. $56.9\%$. This implies that our learned representation, which remains fixed for KNN, but gets ``tuned'' for Fine tuning, generalizes better across action classes than across scene appearance. This is not surprising, since despite neglecting most of the background scene, the \emph{hand-context} box includes some background as well as dataset-dependent color/lighting, whereas incorporating grasp and motion cues strengthens the model's discriminative ability across actions.
We observe that I3D features are performing quite poorly with KNN evaluations compared to our approach, as it can be clearly observed from the first column of \tabref{tbl:knn}. This suggests that the proposed visual cues form better clusters in the latent space by extracting useful information from the input sequences.
\section{Conclusion}

We proposed a methodology that does not demand detailed annotation, and utilizes contextual visual cues for effective activity learning. Specifically, we introduced a simple approach to decouple the foreground action from background appearance via hand detection. We leverage a set of cues, including class-agnostic object-object interaction, hand grasp, optical flow, and hand trajectory to train action RNNs, that we demonstrate to have superior transferability than state-of-the-art action models. We further propose Attentive MAML, an algorithm combining MAML with an attention mechanism for effective transfer learning.
We believe that our inter-class and inter-dataset transfer learning results represent a step-forward in generalization across significant environment and object appearances.

\label{sec:conclusion}

%



\ifCLASSOPTIONcompsoc
  \section*{Acknowledgments}
\else
  \section*{Acknowledgment}
\fi

The authors would like to thank David Joseph Tan for the valuable discussions and constructive feedback. This work was
supported by Microsoft.

\ifCLASSOPTIONcaptionsoff
  \newpage
\fi



%

{\small
\bibliographystyle{IEEEtran}
\bibliography{bibliography}
}
\vspace{-5mm}

\begin{IEEEbiography}[{\includegraphics[width=1.1in,height=1.5in,clip,keepaspectratio]{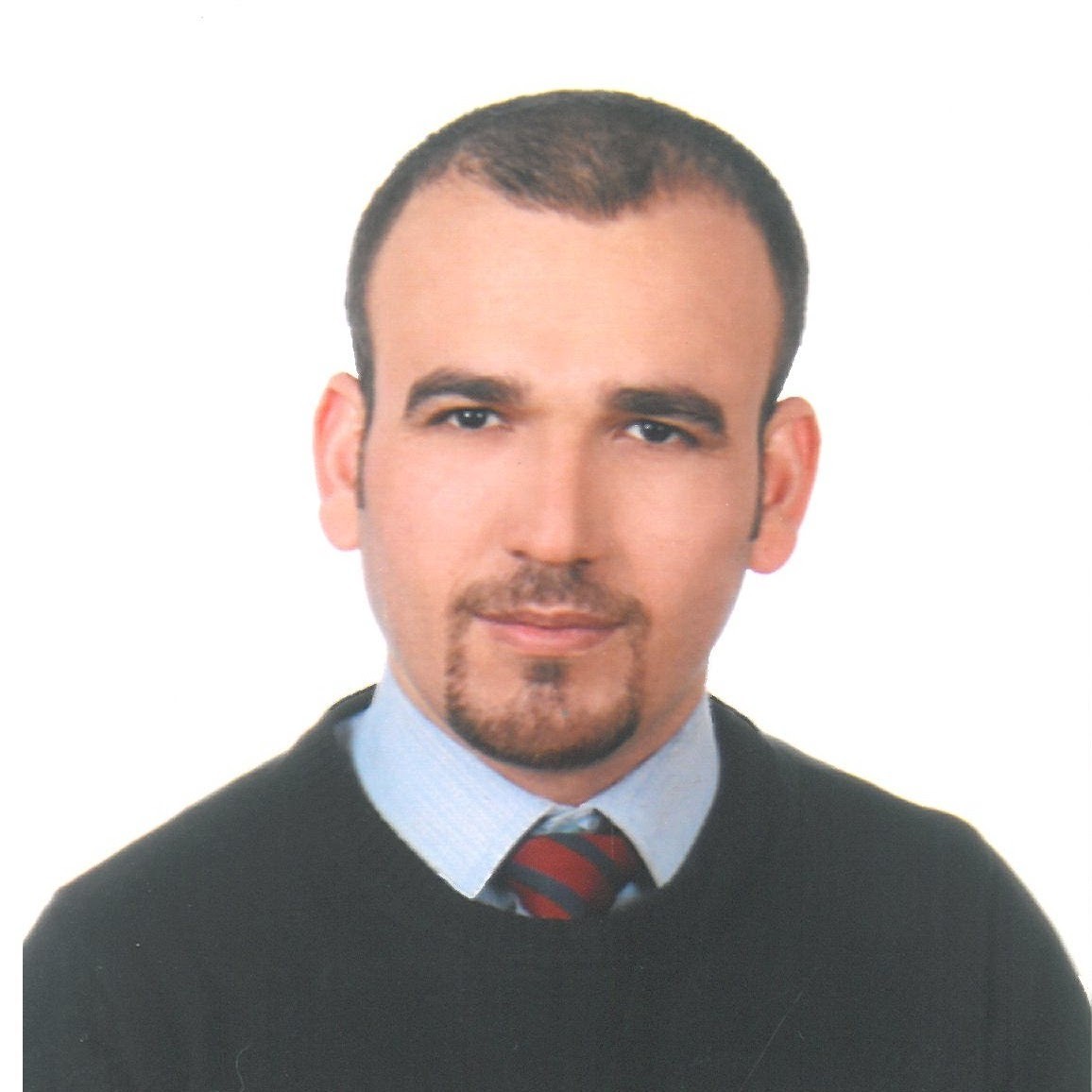}}]{Huseyin Coskun} is a PhD candidate at the Computer Aided Medical Procedures and Augmented Reality (CAMP) laboratories at Technische Universit\"at M\"unchen. He obtained his B. Sc. degree in Mathematics from Instanbul Technical University in 2011, and his M.Sc. degree in Artificial Intelligence from Polytechnic University of Catalonia. He interned two times in Hololens/Vision at Microsoft. His work focuses on activity recognition, few-shot learning and meta learning.

\end{IEEEbiography}
\vspace{-5mm}
\begin{IEEEbiography}[{\includegraphics[width=1.1in,height=1.5in,clip,keepaspectratio]{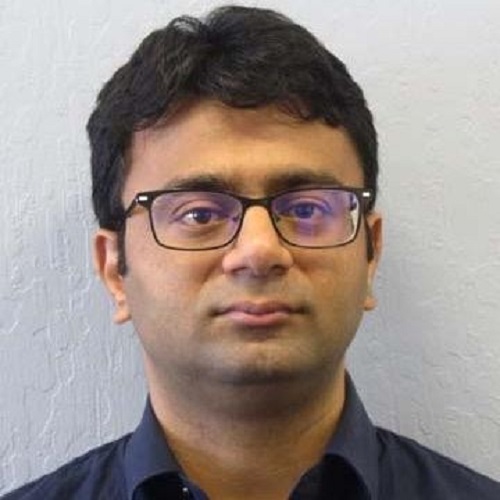}}]{Zeeshan Zia} is a co-founder at Retrocausal working on computer vision problems for the enterprise. Earlier, he pursued research stints at Microsoft, NEC Labs America, and Qualcomm Research where he worked on activity understanding and 3D object localization. He held a postdoctoral fellowship at Imperial College London (2014-15), and studied for a PhD at the Swiss Federal Institute of Technology, Zurich (2009-13), researching the intersection of semantics and geometry in computer vision. He completed his undergraduate education at the Munich University of Technology in Electrical Engineering with a focus on robot vision.
\end{IEEEbiography}
\vspace{-5mm}
\begin{IEEEbiography}[{\includegraphics[width=1.1in,height=1.5in,clip,keepaspectratio]{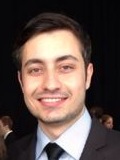}}]{Bugra Tekin} is a researcher at the Microsoft Z\"urich, where he works on computer vision and machine learning topics for mixed reality applications. He received his PhD degree at the Computer Vision Laboratory at \'Ecole Polytechnique F\'ed\'erale de Lausanne (EPFL). Before that, he obtained his M.Sc. degree from EPFL in 2013, and B.Sc degree from Bogazici University in 2011 with high honors. He also spent time at Microsoft Research during his Ph.D. He is the recipient of Qualcomm Innovation Fellowship Europe in 2017. His work focuses on human pose estimation, hand pose estimation, action recognition, 3D object detection and 6D pose estimation.
\end{IEEEbiography}

\begin{IEEEbiography}[{\includegraphics[width=1.1in,height=1.5in,clip,keepaspectratio]{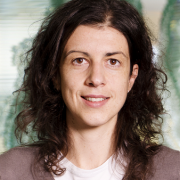}}]{Federica Bogo} is a senior scientist at Microsoft, where she works on computer vision and machine learning topics for mixed-reality applications. She received her PhD in Computer Vision in 2015 from the University of Padova (Italy). From 2012 to 2015, she was an affiliated Ph. D. student at the Max Planck Institute for Intelligent Systems in Tuebingen (Germany), where she also spent one year as postdoctoral researcher (2015-2016). Part of the technology she developed during her Ph. D. and postdoc was licensed to BodyLabs Inc., a startup commercializing body shape technology acquired by Amazon in 2017. In 2016 she joined Microsoft, where she contributed to the development of the hand tracking technology shipped with HoloLens 2.
\end{IEEEbiography}

\begin{IEEEbiography}[{\includegraphics[width=1.1in,height=1.5in,clip,keepaspectratio]{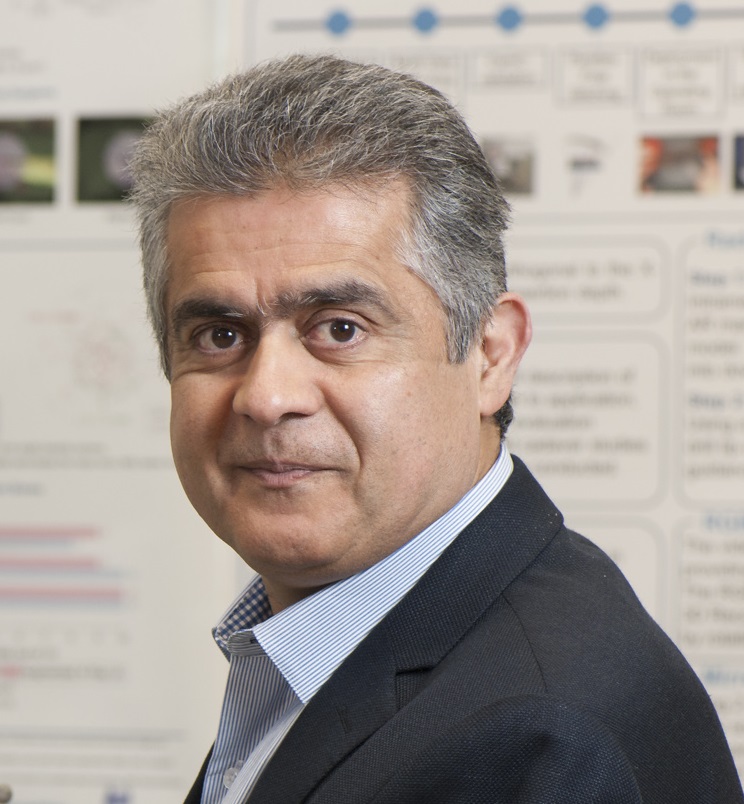}}]{Nassir Navab} is a full Professor and Director of the Laboratory for Computer Aided Medical Procedures, Technical University of Munich and Johns Hopkins University. He has also secondary faculty appointments at both affiliated Medical Schools. He completed his PhD at INRIA and University of Paris XI, France, and enjoyed two years of post-doctoral fellowship at MIT Media Laboratory before joining Siemens Corporate Research (SCR) in 1994. At SCR, he was a distinguished member and received the Siemens Inventor of the Year Award in 2001. He received the SMIT Society Technology award in 2010 and the ‘10 years lasting impact award’ of IEEE ISMAR in 2015. In 2012, he was elected as a Fellow of the MICCAI Society. He has acted as a member of the board of directors of the MICCAI Society and serves on the Steering committee of the IEEE Symposium on Mixed and Augmented Reality (ISMAR) and Information Processing in Computer Assisted Interventions (IPCAI). He is the author of hundreds of peer reviewed scientific papers, with more than 38700 citations and an h-index of 89 as of January 2020. He is the inventor of 50 granted US patents and more than 50 International ones. His current research interests include medical augmented reality, computer-aided surgery, medical robotics, and machine learning.
\end{IEEEbiography}

\begin{IEEEbiography}[{\includegraphics[width=1.1in,height=1.5in,clip,keepaspectratio]{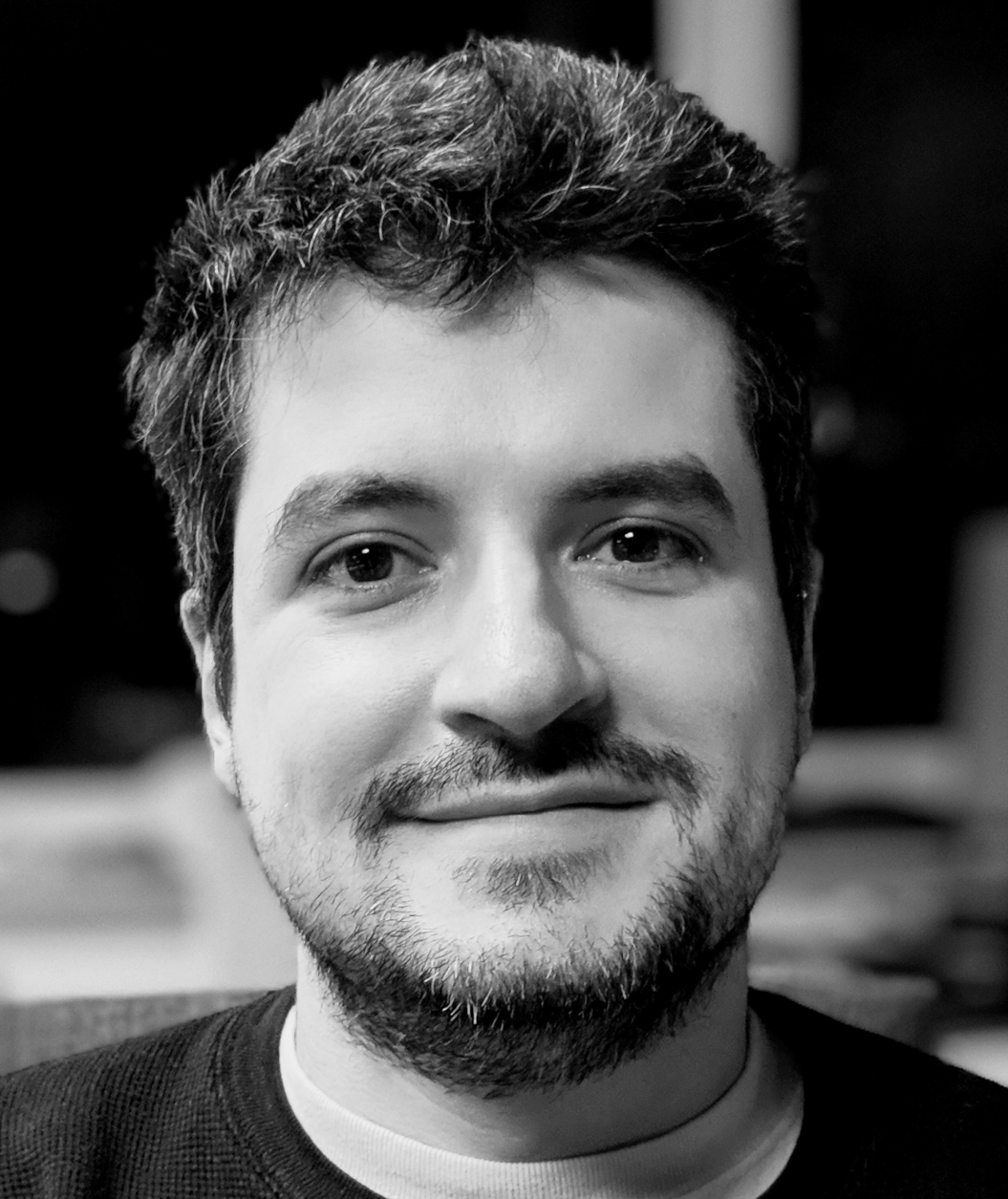}}]{Federico Tombari} is a Research Scientist and Manager at Google, where he leads an applied research team in computer vision and machine learning. Since 2014 he has been Senior Research Scientist first, currently Lecturer (PrivatDozent) at the Technical University of Munich. He has 200+ peer-reviewed publications in the field of computer vision and machine learning and their applications to robotics, autonomous driving, healthcare and augmented reality. He got his PhD in 2009 from the University of Bologna, where he was Assistant Professor from 2013 to 2016. In 2018-19 he was co-founder and managing director of a Munich-based startup on 3D perception for AR and robotics. He regularly serves as Chair and AE for international conferences and journals (ECCV18, 3DV19, ICMVA19, 3DV20, IROS20, ICRA20, RA-L among others). He was the recipient of two Google Faculty Research Awards (in 2015 and 2018), an Amazon Research Award (in 2017), 2 CVPR Outstanding Reviewer Awards (2017, 2018). His works have been awarded at conferences and workshops such as 3DIMPVT'11, MICCAI'15, ECCV-R6D'16, AE-CAI'16, ISMAR '17.
\end{IEEEbiography}

\begin{IEEEbiography}[{\includegraphics[width=1.1in,height=1.5in,clip,keepaspectratio]{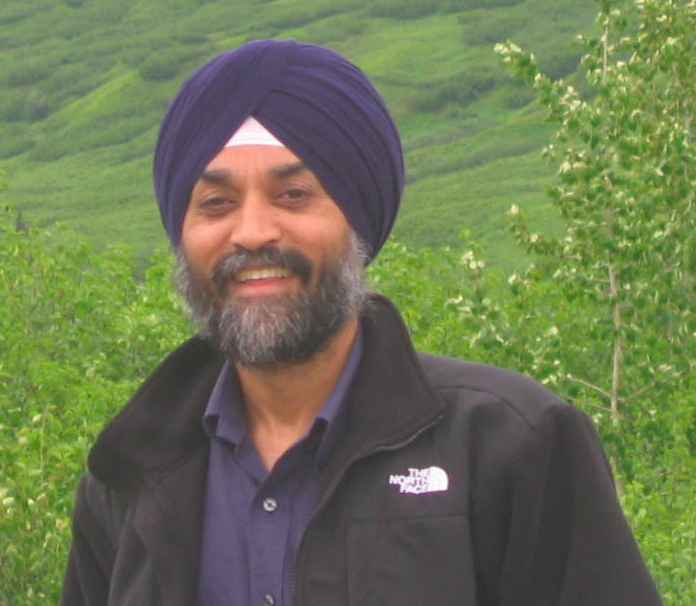}}]{Harpreet S. Sawhney} is a Principal Architect for 3D Vision \& Learning and Mixed Reality in Hololens/Vision at Microsoft.
His passion is to change the way people work and play by augmenting humans through visual understanding of scenes, objects and humans. As CTO-Vision Technologies at SRI (Sarnoff Corp.) in Princeton, NJ, Harpreet led high impact
projects for DARPA and IARPA. These include DARPA’s Visual Media Reasoning (VMR), Exploitation of 3D Data (E3D), Combat Zones that See (CZTS), and IARPA’s pioneering video event detection program ALADDIN. He was the Tech Lead
for two Sarnoff startups, VideoBrush and LifeClips. Harpreet obtained his PhD in Computer Vision from University of Massachusetts, Amherst. He led IBM’s Video QBIC (Query by Image Content) system. He has published over 110 papers
in reputed Computer Vision journals and conferences. He is an inventor on 85 US patents. He served as the Associate Editor of IEEE T-PAMI and as an Area Chair for Computer Vision conferences. Harpreet was awarded an IEEE Fellow
award in 2012 and was honored with an SRI Fellow award in 2011 for his contributions to Computer Vision.

\end{IEEEbiography}




%







\end{document}